%% file: main.tex
\documentclass[acmtog]{acmart}

\AtBeginDocument{%
  \providecommand\BibTeX{{%
    \normalfont B\kern-0.5em{\scshape i\kern-0.25em b}\kern-0.8em\TeX}}}

\input{macros.tex}

\acmSubmissionID{127}

\usepackage{booktabs} 

\AtBeginDocument{%
  \providecommand\BibTeX{{%
    \normalfont B\kern-0.5em{\scshape i\kern-0.25em b}\kern-0.8em\TeX}}}

\usepackage[ruled]{algorithm2e} 

\SetAlFnt{\small}
\SetAlCapFnt{\small}
\SetAlCapNameFnt{\small}
\SetAlCapHSkip{0pt}

\setcopyright{acmlicensed}
\acmJournal{TOG}
\acmYear{2020}
\acmVolume{39}
\acmNumber{4}
\acmArticle{71}
\acmMonth{7}
\acmDOI{10.1145/3386569.3392377}

\begin{document}
\title{Consistent Video Depth Estimation}

\author{Xuan Luo}
\authornote{This work was done while Xuan was an intern at Facebook.}
\affiliation{
\institution{University of Washington}
}
\author{Jia-Bin Huang}
\affiliation{
\institution{Virginia Tech}
}
\author{Richard Szeliski}
\affiliation{
\institution{Facebook}
}
\author{Kevin Matzen}
\affiliation{
\institution{Facebook}
}
\author{Johannes Kopf}
\affiliation{
\institution{Facebook}
}


\begin{abstract}
\input{tex/abstract.tex}
\end{abstract}

%
%


%
%


\begin{teaserfigure}
\input{figures/teaser.tex}
\end{teaserfigure}

\maketitle

\input{tex/intro.tex}
\input{tex/previous.tex}

\input{tex/method.tex}

\input{tex/results.tex}

\input{tex/conclusions.tex}
\input{tex/acknowledge}
 
\bibliographystyle{ACM-Reference-Format}
\bibliography{bibliography}

\appendix

\end{document}

%% file: macros.tex
 \usepackage{enumitem}
\usepackage{soul}
\usepackage{xspace}
\usepackage{mathtools}
\usepackage{adjustbox}
\usepackage{colortbl}
\usepackage[binary-units=true]{siunitx}
\usepackage{mathtools}
\usepackage{multirow}

\soulregister\xspace{0}

\DeclarePairedDelimiter\floor{\lfloor}{\rfloor}

\definecolor{darkgreen}{rgb}{0,0.4,0}

\newcommand{\xuan}[1]{{\color{magenta}\textbf{Xuan: }#1}\normalfont}

\newcommand{\rev}[1]{#1}

\newcommand{\secref}[1]{Section~\ref{sec:#1}}
\newcommand{\figref}[1]{Figure~\ref{fig:#1}} 
\newcommand{\tabref}[1]{Table~\ref{tab:#1}}

\newcommand{\tb}[1]{\textbf{#1}}

\newcommand{\ignorethis}[1]{}

\newcommand{\mfigure}[2]
{
\includegraphics[width=#1]{#2}
}
\newcommand{\mpage}[2]
{
\begin{minipage}{#1} \centering
#2
\end{minipage}
}

\long\def\ignorethis#1{}

\newcommand{\jbox}[2]{%
  \fbox{%
  	\begin{minipage}[b]{#1}%
  		\hfill\vspace{#2}%
  	\end{minipage}%
  }}

\newbox\jsavebox%
\newcommand{\jsubfig}[2]{%
	\sbox\jsavebox{\vspace{0mm}\centering#1}%
	\parbox[t]{\wd\jsavebox}{\vspace{0mm}\centering\usebox\jsavebox\\#2}%
	}



%% file: tex/abstract.tex
We present an algorithm for reconstructing dense, geometrically consistent depth for all pixels in a monocular video.
We leverage a conventional structure-from-motion reconstruction to establish geometric constraints on pixels in the video.
Unlike the ad-hoc priors in classical reconstruction, we use a learning-based prior, i.e., a convolutional neural network trained for single-image depth estimation.
At test time, we fine-tune this network to satisfy the geometric constraints of a particular input video, while retaining its ability to synthesize plausible depth details in parts of the video that are less constrained.
We show through quantitative validation that our method achieves higher accuracy and a higher degree of geometric consistency than previous monocular reconstruction methods.
Visually, our results appear more stable.
Our algorithm is able to handle challenging hand-held captured input videos with a moderate degree of dynamic motion.
The improved quality of the reconstruction enables several applications, such as scene reconstruction and advanced video-based visual effects.

%% file: figures/teaser.tex
\newlength\height%
\setlength\height{3.5cm}%
\centering%
\jsubfig{\includegraphics[height=\height,trim=66 0 66 0,clip]{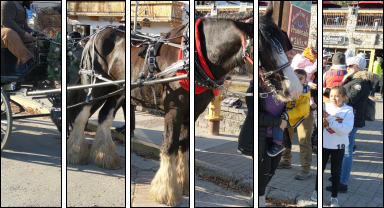}}{\vspace{-1.5mm}%
  {\parbox{0.25\linewidth}{\centering \footnotesize Frame 1}}%
  {\parbox{0.25\linewidth}{\centering \footnotesize Frame 2}}%
  {\parbox{0.25\linewidth}{\centering \footnotesize Frame 3}}%
  {\parbox{0.25\linewidth}{\centering \footnotesize Frame 4}}%
  \\%
  (a) Input video}%
\hfill%
\jsubfig{\includegraphics[height=\height,trim=66 0 66 0,clip]{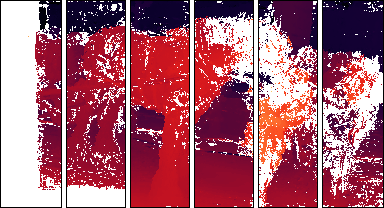}}{\vspace{-1.5mm}%
  {\parbox{0.25\linewidth}{\centering \footnotesize Frame 1}}%
  {\parbox{0.25\linewidth}{\centering \footnotesize Frame 2}}%
  {\parbox{0.25\linewidth}{\centering \footnotesize Frame 3}}%
  {\parbox{0.25\linewidth}{\centering \footnotesize Frame 4}}%
  \\%
  (b) COLMAP depth }%
\hfill%
\jsubfig{\includegraphics[height=\height,trim=66 0 66 0,clip]{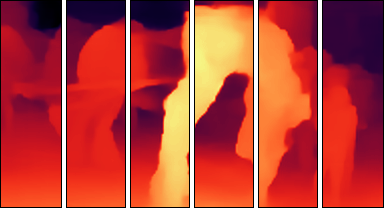}}{\vspace{-1.5mm}%
  {\parbox{0.25\linewidth}{\centering \footnotesize Frame 1}}%
  {\parbox{0.25\linewidth}{\centering \footnotesize Frame 2}}%
  {\parbox{0.25\linewidth}{\centering \footnotesize Frame 3}}%
  {\parbox{0.25\linewidth}{\centering \footnotesize Frame 4}}%
  \\%
  (c) Mannequin Challenge depth} %
\hfill%
\jsubfig{\includegraphics[height=\height,trim=66 0 66 0,clip]{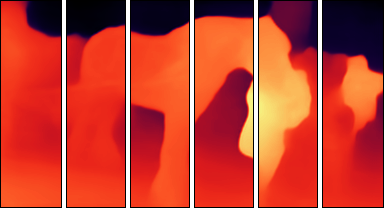}}{\vspace{-1.5mm}%
  {\parbox{0.25\linewidth}{\centering \footnotesize Frame 1}}%
  {\parbox{0.25\linewidth}{\centering \footnotesize Frame 2}}%
  {\parbox{0.25\linewidth}{\centering \footnotesize Frame 3}}%
  {\parbox{0.25\linewidth}{\centering \footnotesize Frame 4}}%
  \\%
  (d) Our result}%
\vspace{-3mm}\\%
\def\teasercaption{
We present a system for estimating temporally coherent and geometrically consistent depth from a casually captured video.
Conventional multi-view stereo methods such as COLMAP \cite{schonberger2016structure} often produce incomplete depth on moving objects or poorly textured areas.
Learning-based methods (e.g., \cite{li2019learning}) predict dense depth for each frame but the video reconstruction is flickering and geometrically inconsistent.
Our video depth estimation is fully dense, globally scale-consistent, and capable of handling dynamically moving objects.
We evaluate our method on a wide variety of challenging videos and show that our results enable new video special effects.
}
\caption{\teasercaption}
\Description[We present a system for consistent video depth estimation.]{\teasercaption}
\label{fig:teaser}
\undef\height

%% file: tex/intro.tex


\section{Introduction}

3D scene reconstruction from image sequences has been studied in our community for decades.
Until a few years ago, the structure from motion systems for solving this problem were not very robust, and practically only worked ``in the lab'', with highly calibrated and predictable setups.
They also, often, produced only sparse reconstructions, i.e., resolving depth at only a few isolated tracked point features.
But in the last decade or so, we have seen good progress towards enabling more \emph{casual} capture and producing \emph{denser} reconstructions, driven by high-quality open-source reconstruction systems and recent advances in learning-based techniques, as discussed in the next section.

Arguably the \emph{easiest} way to capture for 3D reconstruction is using hand-held cell phone video, since these cameras are so readily and widely available, and enable truly spontaneous, impromptu capture, as well as quickly covering large spaces.
If we could achieve fully dense and accurate reconstruction from such input it would be immensely useful---%
however, this turns out to be quite difficult.

Besides the typical problems that any reconstruction system has to deal with, such as poorly textured areas, repetitive patterns, and occlusions, there are several additional challenges with video:
higher noise level,
shake and motion blur,
rolling shutter deformations,
small baseline between adjacent frames,
and, often, the presence of dynamic objects, such as people.
For these reasons, existing methods often suffer from a variety of problems, such as missing regions in the depth maps (Figure~\ref{fig:teaser}b) and inconsistent geometry and flickering depth (Figure~\ref{fig:teaser}c).

Traditional reconstruction methods \cite{Szeliski2010} combine sparse structure-from-motion with dense multi-view stereo---essentially matching patches along epipolar lines.
When the matches are correct, this results in geometrically accurate reconstructions.
However, due to the before-mentioned complications, the matches are often noisy, and typically need to be regularized with heuristic smoothness priors.
This often induces incorrect geometry in the affected regions, so that many methods drop pixels with low confidence altogether, leaving ``holes'' in the reconstruction (Figure~\ref{fig:teaser}b).

There has recently been immense progress on \emph{learning-based} methods that operate on single images.
These methods do not require heuristic regularization, but instead learn scene priors from data, which results in better ability to synthesize plausible depth in parts of the scene that would be weakly or even incorrectly constrained in traditional reconstruction approaches.
They excel, in particular, at the reconstruction of dynamic scenes, since static and dynamic objects appear the same when we consider a single frame at a time.
However, the estimated depth often flickers erratically due to the independent per-frame processing (Figure~\ref{fig:teaser}c), and it is not metric (i.e., not related to true depth by a single scale factor).
This causes a video reconstruction to be \emph{geometrically inconsistent}: objects appear to be attached to the camera and ``swimming'' in world-space.

Several video-based depth estimation methods have also been developed.
These methods address the geometrical consistency of the reconstruction over time either implicitly via recurrent neural networks \cite{wang2019recurrent,Patil2020dont} or explicitly using multi-view reconstruction \cite{liu2019neural,teed2020deepv2d}.
State-of-the-art video-based depth estimation methods \cite{liu2019neural,teed2020deepv2d}, however, handle only static scenes.

In this work, we present a new video-based reconstruction system that combines the strengths of traditional and learning-based techniques.
It uses traditionally-obtained geometric constraints where they are available to achieve accurate and consistent depth, and leverages learning-based priors to fill in the weakly constrained parts of the scene more plausibly than prior heuristics.
Technically, this is implemented by fine-tuning the weights of a single-image depth estimation network at test time,
so that it learns to satisfy the geometry of a particular scene while retaining its ability to synthesize plausible new depth details where necessary.
Our test-time training strategy allows us to use both short-term and long-term constraints and prevent drifting over time.
The resulting depth videos are fully dense and detailed, with sharp object boundaries.
The reconstruction is flicker-free and geometrically consistent throughout the video.
For example, static objects appear rock-steady when projected into world space.
The method even supports a gentle amount of dynamic scene motion, such as hand-waving (Figure~\ref{fig:effects}), although it still breaks down for extreme object motion.

The improved quality and consistency of our depth videos enable interesting new applications, such as fully-automatic video special effects that interact with the dense scene content (Figure~\ref{fig:effects}).
We extensively evaluate our method quantitatively and show numerous qualitative results.
The source code of our method is publicly available.\footnote{ \rev{\texttt{https://roxanneluo.github.io/Consistent-Video-Depth-Estimation/}}}

%% file: tex/previous.tex
\section{Related Work}

\paragraph{Supervised monocular depth estimation.}
Early learning-based approaches regress local image features to depth \cite{saxena2008make3d} or discrete geometric structures \cite{hoiem2005geometric}, followed by some post-processing steps (e.g., a MRF).
Deep learning based models have been successfully applied to single image depth estimation \cite{eigen2014depth,eigen2015predicting,laina2016deeper,liu2015learning,fu2018deep}.
However, training these models requires ground truth depth maps that are difficult to acquire.
Several efforts have been made to address this issue, e.g., training on synthetic dataset \cite{mayer2016large} followed by domain adaptation \cite{atapour2018real}, collecting relative depth annotations \cite{chen2016single}, using conventional structure-from-motion and multi-view stereo algorithms to obtain pseudo ground truth depth maps from Internet images \cite{li2018megadepth,li2019learning,chen2019learning}, or 3D movies \cite{wang2019web,lasinger2019towards}.
Our method builds upon recent advances in single image depth estimation and further improves the geometric consistency of the depth estimation on videos.



\paragraph{Self-supervised monocular depth estimation.} 
Due to challenges of scaling up training data collection, self-supervised learning methods have received considerable attention for their ability to learn a monocular depth estimation model directly from raw stereo pairs \cite{godard2017unsupervised} or monocular video \cite{zhou2017unsupervised}.
The core idea is to apply differentiable warp and minimize photometric reprojection error.
Recent methods improve the performance through incorporating \emph{coupled training} with optical flow \cite{zou2018df,yin2018geonet,ranjan2019competitive}, object motion \cite{vijayanarasimhan2017sfm,dai2019self}, surface normal \cite{qi2018geonet}, edge \cite{yang2018lego}, and visual odometry \cite{wang2018learning,andraghetti2019enhancing,shi2019self}.
Other notable efforts include using stereo information \cite{guo2018learning,watson2019self}, better network architecture and training loss design \cite{gordon2019depth,guizilini2019packnet}, scale-consistent ego-motion network \cite{bian2019unsupervised}, incorporating 3D geometric constraints \cite{mahjourian2018unsupervised}, and learning from unknown camera intrinsics \cite{gordon2019depth,chen2019self}. 

\rev{
Many of these self-supervised methods use a \emph{photometric} loss.
However, these losses can be satisfied even if the geometry is not consistent (in particular, in poorly textured areas). 
In addition, they do not work well for temporally distant frames because of larger appearance changes. 
In our ablation study, however, we show that long-range temporal constraints are important for achieving good results.
}

\paragraph{Multi-view reconstruction.}
Multi-view stereo algorithms estimate scene depth using multiple images captured from arbitrary viewpoints \cite{seitz2006comparison,furukawa2015multi,schonberger2016structure}.
Recent learning-based methods \cite{ummenhofer2017demon,huang2018deepmvs,yao2018mvsnet,im2019dpsnet,kusupati2019normal} leverage well-established principles in traditional geometry-based approaches (e.g., cost aggregation and plane-sweep volume) and show state-of-the-art performance in multi-view reconstruction.
However, these multi-view stereo techniques assume a \emph{static} scene.
For dynamic objects, these methods either produce erroneous estimates or drop pixels with low confidence.
In contrast, our method produces dense depth even in the presence of moderate dynamic scene motion.

\paragraph{Depth from video.} 
Recovering dense depth from monocular video is a challenging problem. 
To handle moving objects, existing techniques rely on motion segmentation and explicit motion modeling for the moving objects in the scene \cite{karsch2014depth,ranftl2016dense,casser2019depth}.
Several methods estimate depth by integrating motion estimation and multi-view reconstruction using two frames \cite{ummenhofer2017demon,wang2019web} or a varying number of frames \cite{zhou2018deeptam,bloesch2018codeslam,valentin2018depth}.
The state-of-the-art video-to-depth methods \cite{teed2020deepv2d,liu2019neural} regress depth (or predict a distribution over depth) based on the cost volume constructed by warping nearby frames to a reference viewpoint.
Such model designs thus do not account for dynamically moving objects. 
In contrast, while we also leverage constraints derived from multi-view geometry, our depth is estimated from (fine-tuned) \emph{single-image} depth estimation models, and thereby handle dynamic object naturally and without the need for explicit motion segmentation.







\paragraph{Temporal consistency.}
Applying single-image based methods independently to each frame in a video often produce flickering results. 
In light of this, various approaches for enforcing temporal consistency have been developed in the context of style transfer \cite{chen2017coherent,ruder2016artistic,huang2017real}, image-based graphics applications \cite{lang2012practical}, video-to-video synthesis \cite{wang2018video}, or application-agnostic post-processing algorithms \cite{bonneel2015blind,lai2018learning}.
The core idea behind these methods is to introduce a ``temporal consistency loss" (either at training or testing time) that encourages similar values along the temporal correspondences estimated from the input video.
In the context of depth estimation from video, several efforts have been made to make the estimated depth more temporally consistent by explicitly applying optical flow-based consistency loss \cite{karsch2014depth} or implicitly encouraging temporal consistency using recurrent neural networks \cite{zhang2019exploiting,wang2019recurrent,Patil2020dont}.
Our work differs in that we aim to produce depth estimates from a video that are \emph{geometrically} consistent.
This is particularly important for casually captured videos because the actual depth may \emph{not} be temporally consistent due to camera motion over time.

\paragraph{Depth-aware visual effects.} 
Dense depth estimation facilitates a wide variety of visual effects such as synthetic depth-of-field \cite{wadhwa2018synthetic}, novel view synthesis \cite{hedman2017casual,hedman2018instant,hedman2018deep,shih20203D}, and occlusion-aware augmented reality \cite{holynski2018fast}.
Our work on consistent depth estimation from causally captured videos enables several new video special effects.

\paragraph{Test-time training.}
Learning on testing data has been used in several different problem contexts: online update in visual tracking \cite{kalal2011tracking,ross2008incremental}, adapting object detectors from images to videos \cite{tang2012shifting,jain2011online}, and learning video-specific features for person re-identification~\cite{cinbis2011unsupervised,zhang2019tracking}.
The work most closely related to ours is that of \cite{casser2019depth,chen2019self} where they improve monocular depth estimation results by fine-tuning a pre-trained model using the testing video sequence.
\rev{Note that any self-supervised method can be trained at test time (as in \cite{chen2019self,casser2019depth}). 
However, the focus of previous methods is largely on achieving \emph{per-frame accuracy}, while our focus is on achieving an accurate prediction with \emph{global geometric consistency}.
Our method achieves accurate and detailed reconstructions with a higher level of temporal smoothness than previous methods, which is important for many video-based applications.

Aside from these goals, there are important technical differences between our method and prior ones.
The method in \cite{casser2019depth} performs a binary object-level segmentation and estimates rigid per-object transformations.
This is appropriate for rigid objects such as cars in a street scene, but less so for highly deformable subjects such as people. 
The method in \cite{chen2019self} uses a geometric loss, similar to ours. 
However, they only train on consecutive frame pairs and relative poses. 
We use absolute poses and long-term temporal connections, which our ablation shows is critical for achieving good results (\figref{visual-ablation}).
}

%% file: tex/method.tex
\section{Overview}
\label{sec:overview}

\def\Di{D_{\!i}}
\def\DiNN{\Di^\textit{NN}}
\def\DiMVS{\Di^\textit{MVS}}
\def\ij{{i \rightarrow j}}
\def\ji{{j \rightarrow i}}
\def\Lgeo{\mathcal{L}_\ij}
\def\Lrepro{\mathcal{L}_\ij^\textit{spatial}}
\def\Ldisp{\mathcal{L}_\ij^\textit{disparity}}

\input{figures/overview.tex}

Our method takes a monocular video as input and estimates a camera pose as well as a dense, \emph{geometrically consistent} depth map (up to scale ambiguity) for each video frame.
The term geometric consistency not only implies that the depth maps do not flicker over time but also, that all the depth maps are in mutual agreement.
That is, we may project pixels via their depth and camera pose accurately amongst frames.
For example, all observations of a static point should be mapped to a single common 3D point in the world coordinate system without drifting.

Casually captured input videos exhibit many characteristics that are challenging for depth reconstruction.
Because they are often captured with a handheld, uncalibrated camera, the videos suffer from motion blur and rolling shutter deformations.
The poor lighting conditions may cause increased noise level and additional blur. 
Finally, these videos usually contain dynamically moving objects, such as people and animals, thereby breaking the core assumption of many reconstruction systems designed for static scenes.

As we explained in the previous sections, in problematic parts of a scene, traditional reconstruction methods  typically produce ``holes'' (or, if forced to return a result, estimate very noisy depth.)
In areas where these methods are confident enough to return a result, however, it is typically fairly accurate and consistent, because they rely strongly on geometric constraints.

Recent learning-based methods \cite{liu2019neural,lasinger2019towards} have complementary characteristics.
These methods handle the challenges described above just fine because they leverage a strong data-driven prior to predict \emph{plausible} depth maps from any input image.
However, applying these methods independently for each frame results in geometrically inconsistent and temporally flickering results over time.

Our idea is to combine the strengths of both types of methods.
We leverage \rev{existing single-image depth estimation networks~\cite{li2019learning,godard2019digging,lasinger2019towards} that have been trained} to synthesize plausible (but not consistent) depth for general color images, 
and we fine-tune the network using the extracted geometric constraints from a video using traditional reconstruction methods.
The network thus learns to produce geometrically consistent depth on a particular video.


Our method proceeds in two stages:

    \tb{Pre-processing (Section~\ref{sec:preproc}):}
As a foundation for extracting geometric constraints among video frames, we first perform a traditional Structure-from-Motion (SfM) reconstruction pipeline using an off-the-shelf open-source software COLMAP~\cite{schonberger2016structure}. 
To improve pose estimation for videos with dynamic motion, we apply Mask R-CNN \cite{he2017mask} to obtain people segmentation and remove these regions for more reliable keypoint extraction and matching, since people account for the majority of dynamic motion in our videos.
This step provides us with accurate intrinsic and extrinsic camera parameters as well as a sparse point cloud reconstruction.
We also estimate dense correspondence between pairs of frames using optical flow. 
The camera calibration and dense correspondence, together, enable us to formulate our geometric losses, as described below.

The second role of the SfM reconstruction is to provide us with the scale of the scene.
Because our method works with monocular input, the reconstruction is ambiguous up to scale. 
The output of the learning-based depth estimation network is scale-invariant as well.
Consequently, to limit the amount the network has to change, we adjust the scale of the SfM reconstruction so that it matches the learning-based method in a robust average sense.

    \tb{Test-time Training (Section~\ref{sec:finetune}):}
In this stage, which comprises our primary contribution, we fine-tune a pre-trained depth estimation network so that it produces more geometrically consistent depth for a \emph{particular} input video.
In each iteration, we sample a pair of frames and estimate depth maps using the current network parameters (Figure~\ref{fig:overview}).
By comparing the dense correspondence with reprojections obtained using the current depth estimates, we can validate whether the depth maps are geometrically consistent.
To this end, we evaluate two geometric losses, 1) spatial loss and 2) disparity loss and back-propagate the errors to update the network weights (which are shared across for all frames).
Over time, iteratively sampling many frame pairs, the losses are driven down, and the network learns to estimate depth that is geometrically consistent for this video while retaining its ability to provide plausible regularization in less constrained parts.

The improvement is often dramatic, our final depth maps are geometrically consistent, temporally coherent across the entire video while accurately delineate clear occluding boundaries even for dynamically moving objects. 
With depth computed, we can have proper depth edge for occlusion effect and make the geometry of the real scene interact with the virtual objects. 
We show various compelling visual effects made possible by our video depth estimation in Section ~\ref{sec:effect}.


\section{Pre-processing}
\label{sec:preproc}

\paragraph{Camera registration.}
We use the structure-from-motion and multi-view stereo reconstruction software COLMAP \cite{schonberger2016structure} to estimate for each frame $i$ of the $N$ video frames the intrinsic camera parameters $K_i$, the extrinsic camera parameters $(R_i, t_i)$, as well as a semi-dense depth map $\DiMVS$.
We set the values to zeros for pixels where the depth is not defined.

Because dynamic objects often cause errors in the reconstruction, we apply Mask~R-CNN~\cite{he2017mask} to segment out people (the most common ``dynamic objects'' in our videos) in every frame independently, and suppress feature extraction in these areas (COLMAP provides this option).
Since smartphone cameras are typically not distorted\footnote{\rev{Our test sequences (Section~\ref{sec:experimental_setup}) are captured with a fisheye camera, and we remove the distortion through rectification.}}, we use the \texttt{SIMPLE\_PINHOLE} camera model and solve for the shared camera intrinsics for all the frames, as this provides a faster and more robust reconstruction.
We use the exhaustive matcher and enable guided matching.


\paragraph{Scale calibration.}
The scale of the SfM and the learning-based reconstructions typically do \emph{not} match, because both methods are scale-invariant.
This manifests in different value ranges of depth maps produced by both methods.
To make the scales compatible with the geometric losses, we adjust the SfM scale, because we can simply do so by multiplying all camera translations by a factor.

Specifically, let $\DiNN$ be the initial depth map produced by the learning-based depth estimation method. 
We first compute the relative scale for image $i$ as:
\begin{equation}
s_i = \operatornamewithlimits{median}_x\!
\left\{
  \DiNN(x) ~/~ \DiMVS(x)
  ~\middle\vert~
  \DiMVS(x) \neq 0
\right\}\!,
\end{equation}
where $D(x)$ is the depth at pixel $x$.

We can then compute the global scale adjustment factor $s$ as
\begin{equation}
s = \operatornamewithlimits{mean}_i \!\left\{ s_i \right\},
\end{equation}
and update all the camera translations
\begin{equation}
\tilde{t_i} = s \cdot t_i.
\end{equation}


\paragraph{Frame sampling.}
In the next step, we compute a dense optical flow for certain pairs of frames.
This step would be prohibitively computationally expensive to perform for all $O(N^2)$ pairs of frames in the video.
We, therefore, use a simple hierarchical scheme to prune the set of frame pairs down to $O(N)$.

The first level of the hierarchy contains all consecutive frame pairs,
\begin{equation}
S_0 = \big\{~
(i, j) \ \ \big\vert~
\left| i - j \right| = 1
~\big\}.
\end{equation}

Higher levels contain a progressively sparser sampling of frames,
\begin{equation}
S_l = \big\{~
(i, j) \ \ \big\vert~
\left| i - j \right| = 2^l,~
i\,\operatorname{mod}\,2^{l-1}=0
~\big\}.
\end{equation}

The final set of sampled frames is the union of the pairs from all levels,
\begin{equation}
S = \bigcup_{ 0 \leq l \leq \floor{\log_2(N - 1)} } S_l.
\end{equation}

\paragraph{Optical flow estimation.}
For all frame pairs $(i, j)$ in $S$ we need to compute a dense optical flow field $F_\ij$.
Because flow estimation works best when the frame pairs align as much as possible, we first align the (potentially distant) frames using a homography-warp (computed with a RANSAC-based fitting method \cite{Szeliski2010}) to eliminate dominant motion between the two frames (e.g., due to camera rotation).
We then use FlowNet2 \cite{ilg2017flownet} to compute the optical flow between the aligned frames.
To account of moving objects and occlusion/dis-occlusion (as they do not satisfy the geometric contraints or are unreliable), we apply a forward-backward consistency check and remove pixels that have forward-backward errors larger than 1 pixel, producing a binary map $M_\ij$.
Furthermore, we observe that the flow estimation results are not reliable for frame pairs with little \emph{overlap}. 
We thus exclude any frame pairs where $|M_\ij|$ is less than 20\% of the image area from consideration.


\section{Test-Time Training on Input Video}
\label{sec:finetune}

Now we are ready to describe our test-time training procedure, i.e., how we coerce the depth network through fine-tuning it with a geometric consistency loss to producing more consistent depth for a particular input video.
We first describe our geometric loss, and then the overall optimization procedure.


\paragraph{Geometric loss.}
For a given frame pair $(i, j) \in S$, the optical flow field $F_\ij$ describes which pixel pairs show the same scene point.
We can use the flow to test the geometric consistency of our current depth estimates:
if the flow is correct and a flow-displaced point $f_\ij(x)$ is identical to the depth-reprojected point $p_\ij(x)$ (both terms defined below), then the depth \emph{must} be consistent.

The idea of our method is that we can turn this into a geometric loss $\Lgeo$ and back-propagate any consistency errors through the network, so as to coerce it into to producing depth that is \emph{more} consistent than before.
$\Lgeo$ comprises two terms, an image-space loss $\Lrepro$, and a disparity loss $\Ldisp$.
To define them, we first discuss some notation.

Let $x$ be a 2D pixel coordinate in frame $i$.
The flow-displaced point is simply
\begin{equation}
f_\ij(x) = x + F_\ij(x).
\end{equation}

To compute the depth-reprojected point $p_\ij(x)$, we first lift the 2D coordinate to a 3D point $c_i(x)$ in frame $i$'s camera coordinate system, using the camera intrinsics $K_i$ as well as the current depth estimate $D_i$,
\begin{equation}
c_i(x) = D_i(x) \, K_i^{-1} \tilde{x},
\end{equation}
where $\tilde{x}$ is the homogeneous augmentation of $x$.
We then further project the point to the other frame $j$'s camera coordinate system,
\begin{equation}
c_\ij(x) = R_j^\mathsf{T}\!\,\Big( R_i \, c_i(x) + \tilde{t_i} - \tilde{t_j} \Big),
\end{equation}
and finally convert it back to a pixel position in frame $j$,
\begin{equation}
p_\ij(x) = \pi\!\,\big( K_j \, c_\ij(x) \big),
\end{equation}
where $\pi \big( [x, y, z]^\mathsf{T} \big) = [\tfrac{x\vphantom{y}}{z}, \tfrac{y}{z}]^\mathsf{T}$.

With this notation, the image-space loss for a pixel can be easily defined:
\begin{equation}
\Lrepro(x) =
\left\| p_\ij(x) - f_\ij(x) \right\|_2,
\label{eq:loss}
\end{equation}
which penalizes the image-space distance between the flow-displaced and the depth-reprojected point.

The disparity loss, similarly, penalizes the disparity distance in camera coordinate system:
\begin{equation}
\Ldisp(x) =
u_i \left| \, z_\ij^{-1}(x) - z_j^{-1} \!\big( f_\ij(x) \big) \, \right|,
\end{equation}
where $u_i$ is frame $i$'s focal length, and $z_i$ and $z_\ij$ are the scalar z-component from $c_i$ and $c_\ij$, respectively.

The overall loss for the pair is then simply a combination of both losses for all pixels where the flow is valid,
\begin{equation}
\mathcal{L}_\ij =
\frac{1}{\left| M_\ij \right|}
\sum\limits_{x \in M_\ij}
\mathcal{L}_\ij^\textit{spatial}(x) +
\lambda \mathcal{L}_\ij^\textit{disparity}(x),
\end{equation}
where $\lambda = 0.1$ is a balancing coefficient. 


\rev{
\paragraph{Discussion.}
While the second term in Equation~\ref{eq:loss} (flow mapping) can handle dynamic motion, the first term (depth reprojection) assumes a \emph{static} scene.
How can this still result in an accurate depth estimation?
There are two cases:
(1) Consistent motion (e.g., a moving car) can sometimes be aligned with the epipolar geometry and cause our method, like most others, to estimate the wrong depth.
(2) Consistent motion that is \emph{not} epipolar-aligned or inconsistent motion (e.g., a waving hand) causes conflicting constraints;
empirically, our test-time training is tolerant to these conflicting constraints and produces accurate results (as seen in many examples in this submission and accompanying materials.)
}


\paragraph{Optimization.}
Using the geometric loss between $i$-th and $j$-th frames $\mathcal{L}_\ij$, we fine-tune the network weights using standard backpropagation.
Initializing the network parameters using a pre-trained depth estimation model allows us to transfer the knowledge for producing plausible depth maps on images that are challenging for traditional geometry-based reconstructon systems.
We fine-tune the network using a fixed number of epochs (20 epochs for all our experiments).
\rev{
In practice, we find that with this simple fine-tuning step the network training does not overfit the data in the sense that it does not lose its ability to synthesize plausible depth in unconstrained or weakly constrained parts of the scene}\footnote{We note that there are more advanced regularization techniques for transfer learning \cite{li2018explicit,kirkpatrick2017overcoming}. These can be applied to further improve the performance of our method.}.
We also observe that the training handles a certain amount of erroneous supervision (e.g., when the correspondences are incorrectly established). 






\paragraph{Implementation details.}
\rev{We have experimented with several monocular depth estimation architectures and pre-trained weights~\cite{li2019learning,godard2019digging,lasinger2019towards}.
If not otherwise noted, results in the paper and accompanying materials use Li et al.'s network~\shortcite{li2019learning} (single-image model).
We use the other networks in evaluations as noted there.}
Given an input video, an epoch is defined by one pass over all frame pairs in $\mathcal{S}$.
In all of our experiments, we fine-tune the network for 20 epochs with a batch size of 4 and a learning rate of 0.0004 using ADAM optimizer~\cite{kingma2014adam}.
The time for test-time training varies for videos of different lengths. 
\rev{For a video of 244 frames, training on 4 NVIDIA Tesla M40 GPUs takes $40$ min.}

%% file: figures/overview.tex
\begin{figure*}[ht]
    \centering
    \includegraphics[width=1\linewidth]{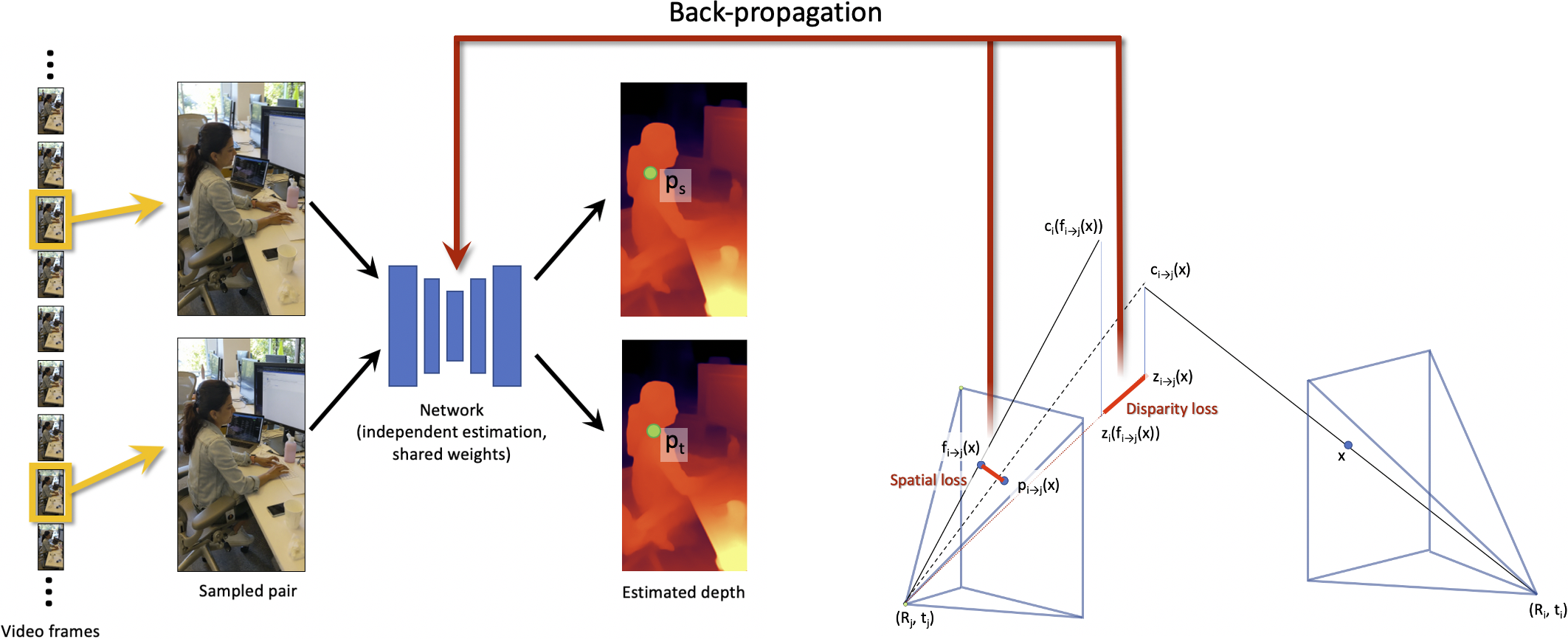}
    \vspace{-2em}
    \caption{
    Method overview. 
    With a monocular video as input, we sample a pair of (potentially distant) frames and estimate the depth using a pre-trained, single-image depth estimation model to obtain initial depth maps. 
    From the pair of images, we establish correspondences using optical flow with forward-backward consistency check. 
    We then use these correspondences and the camera poses to extract geometric constraints in 3D. 
    We decompose the 3D geometric constraints into two losses: 1) spatial loss and 2) disparity loss and use them to fine-tune the weight of the depth estimation network via standard backpropagation. 
    This test-time training enforces the network to minimize the geometric inconsistency error across multiple frames for this particular video.
    After the fine-tuning stage, our final depth estimation results from the video is computed from the fine-tuned model.  
    }
    \label{fig:overview}
    \vspace{-0.5em}
\end{figure*}

%% file: tex/results.tex
\section{Results and Evaluation}
\label{sec:results}

In this section, we first describe the experimental setup (\secref{experimental_setup}).
We then present quantitative comparison with the state-of-the-art depth estimation methods (\secref{evaluation}).
We conduct an extensive ablation study to validate the importance of our design choices and their contributions to the results (\secref{ablation}).
Finally, we show qualitative results of our depth estimation and their applications to new advanced video-based visual effects (\secref{effect}).


\subsection{Experimental Setup} \label{sec:experimental_setup}

\input{figures/test_examples}
\input{figures/quantitative_comparison}

\paragraph{Dataset.}
Many datasets have been constructed for evaluating depth reconstruction.
However, these existing datasets are either for synthetic \cite{mayer2016large,butler2012naturalistic} , specific domains (e.g., driving scenes) \cite{geiger2013vision}, single images \cite{li2018megadepth,li2019learning,chen2016single}, or videos (or multiple images) of static scenes \cite{schops2017multi,silberman2012indoor,sturm2012benchmark}.
Consequently, we capture custom stereo video datasets for evaluation.
Our test set consists of both static and dynamic scenes with a gentle amount of object motion (see Fig.~\ref{fig:test-set} for samples. 
We capture the videos with stereo fisheye QooCam cameras.\footnote{\url{https://www.kandaovr.com/qoocam/}}
\rev{The handheld camera rig provides a handy way to capture stereo videos, but it is highly distorted in the periphery due to the fisheye lenses.}
We, therefore, rectify and crop the center region using the Qoocam Studio\footnote{\url{https://www.kandaovr.com/download/}} and obtain videos of resolution $1920 \times 1440$ pixels. 
The lengths of the captured video range from 119 to 359 frames.
Our new video dataset \rev{is available on the accompanying website} for evaluating future video-based depth estimation.

\rev{For completeness, we also provide quantitative comparisons with the state-of-the-art depth estimation models on three publicly available datasets:
(1) the TUM dataset \cite{sturm12iros} (using the 3D Object Reconstruction category),
(2) the ScanNet dataset \cite{dai2017scannet} (using the testing split provided by \cite{teed2020deepv2d}), and
(3) the KITTI 2015 dataset \cite{geiger2012we} (using the Eigen split \cite{eigen2014depth}).
}


\paragraph{Evaluation metrics.}
To evaluate and compare the quality of the estimated depth from a monocular video on our custom stereo video dataset, we use the following three different metrics.

\input{tables/quantitative_comparison}
\textbf{Photometric error $E_{p}$:}
We use photometric error to quantify the accuracy of the recovered depth. 
All the methods estimate the depth from the \emph{left} video stream. 
Using the estimated depth, we then reproject the pixels from the left video stream to the right one and compute the photometric error as mean squared error of the RGB differences.
As the depth map can only be estimated up to a scale ambiguity, we need to align the estimated depth maps to the stereo disparity. 
Specifically, we compute the stereo disparity by taking the horizontal components from the estimated flow on the stereo pair (using Flownet2~\cite{ilg2017flownet}).
For each video frame, we then compute the \emph{scale} and \emph{shift} alignment to the computed stereo disparity using RANSAC-based linear regression. 
We can obtain the global (video-level) scale and shift parameters by taking the mean of the scales/shifts for all the frames.

\textbf{Instability $E_{s}$:}
We measure instability of the estimated depth maps over time in a video as follows.
We first extract a sparse set of reliable tracks from the input monocular video using a standard KLT tracker.
We then convert the 2D tracks to 3D tracks, using the camera poses and calibrated depths to unproject 2D tracks to 3D.
For a perfectly stable reconstruction, each 3D track should collapse to a single 3D point.
We thus can quantify the instability by computing the Euclidean distances of the 3D points for each pair of consecutive frames.


\textbf{Drift $E_{d}$:}
In many cases, while 3Dtracks described above may appear somewhat stable for consecutive frames, the errors could be accumulated and cause \emph{drift} over time. 
To measure the amount of drift for a particular 3D track, we compute the maximum eigenvalue of the covariance matrix formed by the 3D track. 
Intuitively, this measures how \emph{spread} the 3D points is across time.

For static sequences, we evaluate the estimated depth using all three metrics.
For dynamic sequences, we evaluate only on photometric error and instability, as the drift metric does not account for dynamically moving objects in the scene.

\subsection{Comparative Evaluation}
\label{sec:evaluation}

\paragraph{Compared methods.} 
We compare our results with state-of-the-art depth estimation algorithms from three main categories. 
\begin{itemize}
    \item \tb{Traditional multi-view stereo system}: COLMAP \cite{schonberger2016structure}.
    \item \tb{Single-image depth estimation}: Mannequin Challenge \cite{li2019learning} and MiDaS-v2 \cite{lasinger2019towards}.
    \item \tb{Video-based depth estimation}: WSVD \cite{wang2019web} (two frames) and NeuralRGBD \cite{liu2019neural} (multiple frames).
\end{itemize}

\input{figures/visual_comparison}
\input{figures/visual_ablation}

\paragraph{Quantitative comparison.} 
Fig.~\ref{fig:quantitative-comparison} shows the plot of the photometric error, instability, and drift metrics against  completeness. 
In all three metrics, our method compares favorably against previously published algorithms.
Our results particularly shine when evaluated on the instability and the drift metrics, highlighting the \emph{consistency} of our results. 
\tabref{quantitative-eval} further reports the summary of the results for different methods.

\paragraph{Visual comparison}
We present in Fig.~\ref{fig:visual-comparison} the qualitative comparison of different depth estimation methods. 
The traditional multi-view stereo method produces accurate depths at highly textured regions, where reliable matches can be established. 
These depth maps contain large holes (black pixels), as shown in Fig.~\ref{fig:visual-comparison}b.
The learning-based single-image depth estimation approaches \cite{li2019learning,lasinger2019towards} produce dense, plausible depth maps for each individual video frame. 
However, flickering depths over time cause geometrically inconsistent depth reconstructions.
Video-based methods such as NeuralRGBD alleviate the temporal flicker, yet suffer from drift due to the limited temporal window used for depth estimation. 
We refer the readers to the video results in the supplementary material.



\input{tables/ablation-study}

\input{figures/visual_long_term_ablation}


\subsection{Ablation Study}\label{sec:ablation}
We conduct an ablation study to validate the effectiveness of several design choices in our approaches.
\rev{We first study the effect of losses and the importance of different steps in the pipeline, including scale calibration and overlap test. 
We summarize these results in Table~\ref{tab:ablation}.
Fig.~\ref{fig:visual-ablation} visualizes the contributions of various components. 

We observe that using long-term constraints help improve the stability of the estimated depth over time.
As the disparity loss also helps reduce temporal flickering, we further investigate the effects of both design choices in Fig.~\ref{fig:ablate-long-term}.
Our results show that including constraints from long-term frame pairs leads to sharper and temporally more stable results.
In contrast, while adding disparity loss reduces temporal flickers, it produces blurry results when using only consecutive frame pairs.
}

\rev{
\subsection{Quantitative Comparisons on Public Benchmarks} \label{sec:quan_comparison}

We provide quantitative results on three publicly available benchmark datasets for evaluating the performance of our depth estimation.
In all of the evaluation settings, we resize the input images so that the longest image dimension to 384.
We finetune the monocular depth estimation network for 20 epochs (the same evaluation setting used in the stereo video dataset).



\paragraph{TUM-RGBD dataset.} 
We evaluate our method on the 11 scenes in the ``3D Object Reconstruction" category in the TUM-RGBD dataset \cite{sturm12iros}.
For evaluation, we subsample the videos every 5 frames and obtain sequences ranging from 195 to 593 frames.
Here, we use the ground truth camera pose provided by the dataset.
We then fine-tune the single-image model from Li et al.~\shortcite{li2019learning} on the subsampled frames. 
To compute the error metrics, we align the predicted depth map to the ground truth using per-image median scaling.
We report the errors in the disparity (inverse depth) space as it does not require clipping any depth ranges.

\tabref{quan_TUM} reports the quantitative comparisons with single-frame methods \cite{li2019learning,lasinger2019towards} and multi-frame methods \cite{wang2019web,liu2019neural}. 
Our approach performs favorably against prior methods with a large margin in all evaluation metrics.
In particular, our proposed test-time training significantly improves the performance over the baseline model from Li et al.~\shortcite{li2019learning}.

\paragraph{ScanNet dataset.} 
Following the evaluation protocol of Teed and Deng~\shortcite{teed2020deepv2d}, we evaluate our method on the 2,000 sampled frames from the 90 test scenes in the ScanNet dataset \cite{dai2017scannet}.
We finetune the MiDaS-v2 model \cite{lasinger2019towards} on each testing sequence with a learning rate of $10^{-5}$ and $\lambda=10^{-5}$.
Following Teed and Deng~\shortcite{teed2020deepv2d}, we apply per-image median scaling to align the predicted depth to the ground truth depth map (using the range of [0.5, 8] meters).
We then evaluate all the metrics in the depth space over the regions where the ground truth depth is within the range of 0.1 to 10 meters.

\tabref{quan_ScanNet} shows the quantitative comparisons with several other multi-frame based depth estimation methods \cite{ummenhofer2017demon,tang2018ba,teed2020deepv2d} and the baseline single-image model \cite{lasinger2019towards}. 
Our method achieves competitive performance with the state-of-the-art algorithms, performing slightly inferior to the DeepV2D method that is trained on the ScanNet training set.

\paragraph{KITTI dataset.}
We evaluate our method on the KITTI dataset~\cite{geiger2012we} using the Eigen test split \cite{eigen2014depth} for comparison with prior monocular depth estimation methods.
We estimate the camera poses for each test sequence using COLMAP \cite{schonberger2016structure}.
We observe that the FlowNet2 model (pre-trained on the Flying Chairs~\cite{DFIB15} and Flying Things 3D~\cite{MIFDB16} datasets) performs poorly in the KITTI dataset~\cite{geiger2012we}.
Consequently, we use the FlowNet2 model finetuned on a combination of the KITTI2012~\cite{geiger2012we} and KITTI2015~\cite{Menze2015ISA} training sets, FlowNet2-ft-kitti, for extracting dense correspondence across frames.
Due to the challenging large forward motion in the driving videos, the flow estimations between temporally distant frames are not accurate. 
We thus only sample pairs with more than $50\%$ consistent flow matches.
We use the Monodepth2 \cite{godard2019digging} as the base single-image depth estimation network. We apply our fine-tuning method at the resolution of $384\times 112$ over each sequence with a learning rate of $4\times 10^{-5}$ and $\lambda=1$.
Following the standard protocol \cite{godard2017unsupervised}, we cap the depth to 80m and report the results using the per-image median ground truth scaling alignment. 




\input{figures/quantitative_kitti}

\tabref{quan_KITTI} presents the quantitative comparisons with the state-of-the-art monocular depth estimation methods. 
Under this evaluation setting, the results appear to show that our method does not provide an overall improvement over the baseline model Monodepth2 \cite{godard2019digging}.
To investigate this issue, we show in \figref{quantitative-kitti} the sorted error (Abs Rel) and accuracy ($\sigma < 1.1$) metrics for all the testing frames.
The results show that our method indeed improves the performance in more than 80\% of the testing frames (even when compared with the model with a high-resolution outputs). 
However, as COLMAP produce erroneous pose estimates in sequences with large dynamic objects in the scene, our fine-tuning method inevitably results in depth estimation with very large errors.
Our method also has difficulty in handling significant dynamic scene motion.  
As a result, our method does not achieve clear improvement when the results are averaged over all the testing frames.
Please see the supplementary material for video result comparison.

\input{tables/quantitative_scannet}

\input{tables/quantitative_tum}
\input{tables/quantitative_kitti}

}






\input{figures/effects.tex}
\subsection{Video-based Visual Effects}
\label{sec:effect}
Consistent video depth estimation enables interesting video-based special effects. 
Fig.~\ref{fig:effects} showcases samples of these effects.
Full video results can be found in the supplementary material.

\subsection{Limitations} 
There are several limitations and drawbacks of the proposed video depth estimation method.

\begin{description}
\item[Poses]
Our method currently relies on COLMAP \cite{schonberger2016structure} to estimate the camera pose from a monocular video. 
In challenging scenarios, e.g., limited camera translation and motion blur, however, COLMAP may not be able produce reliable sparse reconstruction and camera pose estimation.
\rev{Large pose errors have a strong degrading effect on our results.}
This limits the applicability of our method on such videos.
Integrating learning-based pose estimation (e.g., as in \cite{liu2019neural,teed2020deepv2d}) with our approach is an interesting future direction. 

\item[Dynamic motion]
Our method supports videos containing moderate object motion. 
It breaks for extreme object motion. 

\rev{
\item[Flow]
We rely on FlowNet2~\cite{ilg2017flownet} to establish geometric constraints.
Unreliable flow is filtered through forward-backward consistency checks, but it might be by chance erroneous in a consistent way.
In this case our method will fail to produce correct depth.
We tried using sparse flow (subsampling dense flow on a regular grid), but it did not perform well.
}

\item[Speed]
As we extract geometric constraints using all the frames in a video, we do not support online processing. 
For example, our test-time training step takes about $40$ minutes for a video of $244$ frames and $708$ sampled flow pairs.
Developing online and fast variants in the future will be important for practical applications.
\end{description}

%% file: figures/test_examples.tex
\begin{figure}[t]
    \setlength{\tabcolsep}{0pt}
    \def\imW{0.24\linewidth}
    \centering
    \begin{tabular}{cc|cc}
         \mfigure{\imW}{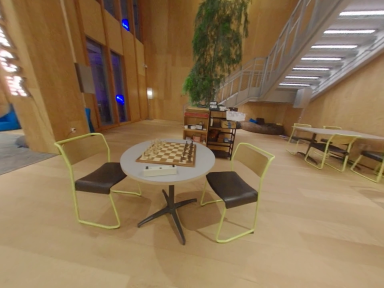}
         & \mfigure{\imW}{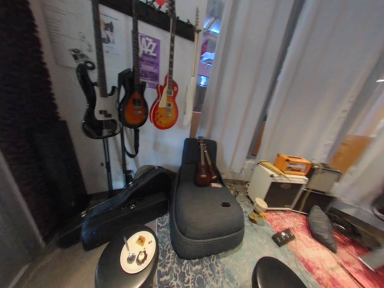}  
         & \mfigure{\imW}{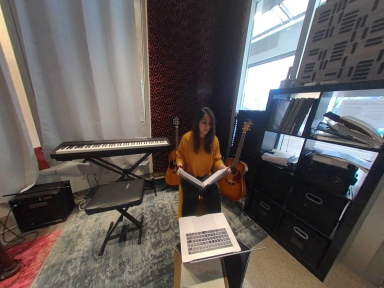}
         &\mfigure{\imW}{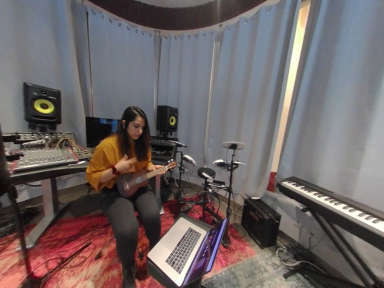} \\
         \mfigure{\imW}{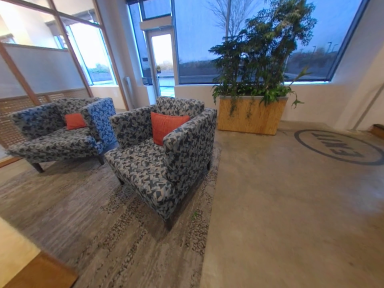}
         & \mfigure{\imW}{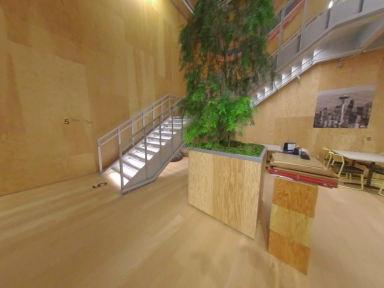}
         & \mfigure{\imW}{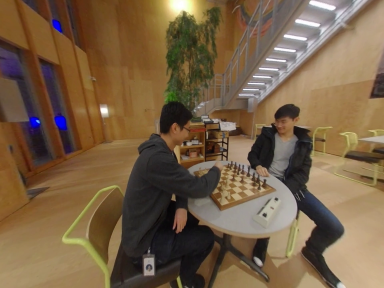} \\
         \multicolumn{2}{c|}{Static} &\multicolumn{2}{c}{Dynamic}
    \end{tabular}
    \vspace{-3mm}
    \caption{Example frames from our test set that includes four static sequences and three dynamic ones. The dynamic videos contain gentle amount of seated motion like playing ukulele and body motion while playing chess, flipping the notes while singing, etc. These videos resemble casual video capture scenario where the hand-held camera is shaky and frames contain motion blur.}
    \label{fig:test-set}
    \vspace{-1em}
\end{figure}

%% file: figures/quantitative_comparison.tex
\begin{figure*}[t]
    \setlength{\tabcolsep}{1pt}
    \def\imW{0.32\linewidth}
    \centering
    \begin{tabular}{ccc}
    \mfigure{\imW}{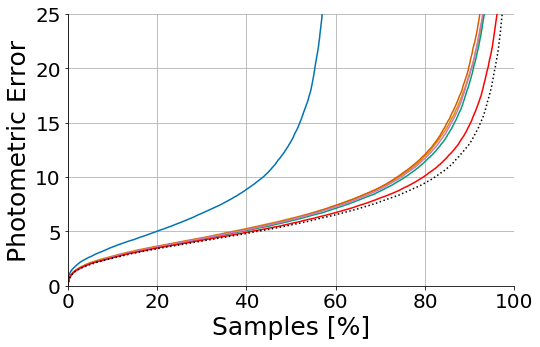} 
    &\mfigure{\imW}{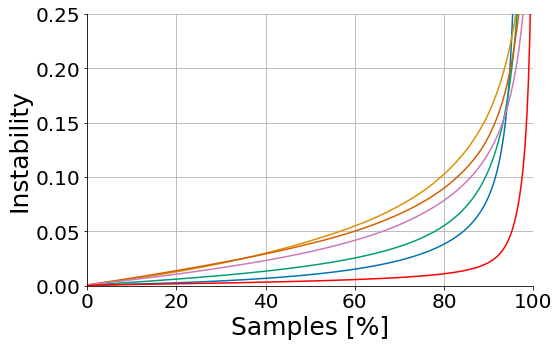} 
    &\mfigure{\imW}{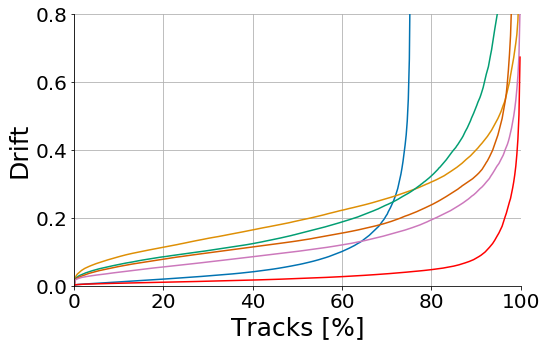} \\
    \mpage{\imW}{Photometric} \hfill
    &\mpage{\imW}{Instability} \hfill
    &\mpage{\imW}{Drift}  \\
    
    \mfigure{\imW}{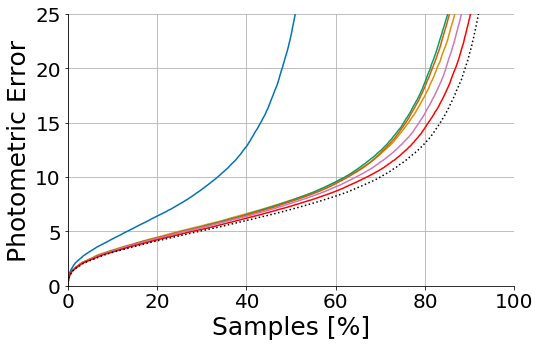}
    &\mfigure{\imW}{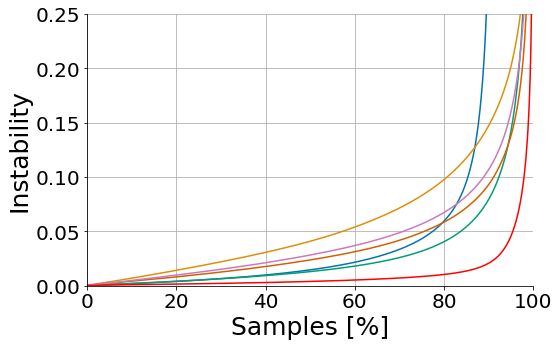}
    &
    \raisebox{5mm}{
    \mfigure{0.2\linewidth}{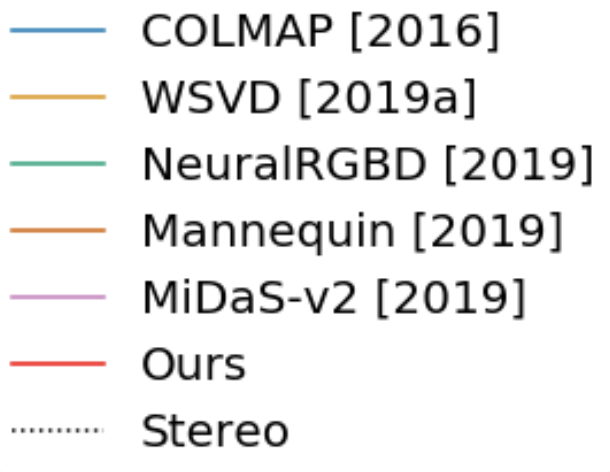} }
    \\
    \mpage{\imW}{Photometric} 
    &\mpage{\imW}{Instability} &\\
    \end{tabular}
    \vspace{-3mm}
    \caption{
Quantitative comparison with the state-of-the-art.
We plot the error of all reconstructed pixels, sorted by error.
\rev{
Note, that COLMAP drops some pixels from the reconstruction.
Hence, its curve in the left column stops short of 100\%;
the second and third column evaluate on tracks, which tend to be in textured areas where COLMAP has a higher level of completeness.
}
Top: static sequences; bottom: Dynamic sequences.
}
    \vspace{-3mm}
    \label{fig:quantitative-comparison}
\end{figure*}

%% file: tables/quantitative_comparison.tex
\begin{table}[t]
    \setlength{\tabcolsep}{2pt}
    \centering
    \caption{Quantitative comparisons with the state-of-the-art depth estimation algorithms.}
    \label{tab:quantitative-eval}
    \begin{tabular}{lcccccc}
    \toprule
    \multirow{2}{*}{} & 
    \multicolumn{3}{c}{Static} &
    \multicolumn{3}{c}{Dynamic} \\
    \cmidrule{2-4} \cmidrule{6-7}
        & $E_s$ (\%) $\downarrow$ & $E_{d}$ (\%) $\downarrow$ & 
        $E_p$ $\downarrow$ & & 
        $E_s$ (\%) $\downarrow$ & $E_p$ $\downarrow$ \\
    \midrule
    WSVD \shortcite{wang2019web}                    &4.13 &19.12 &11.90 &&4.10 & 17.46 \\
    NeuralRGBD \shortcite{liu2019neural}            &1.86 &15.25 &11.33 &&1.30 & 18.62\\
    Mannequin \shortcite{li2019learning}            &3.88 &13.22 &12.05 &&2.38 & 18.16\\
    MiDaS-v2 \shortcite{li2019learning}             &3.14 &10.14 &11.74 &&2.83 & 15.76\\
    COLMAP \shortcite{schonberger2016structure}     &1.02 &6.19 & -     &&1.47 & -\\
    Ours                                            &\textbf{0.44} &\textbf{2.12} &  \textbf{10.09}  &&\textbf{0.40} & \textbf{14.44}\\
    \bottomrule
    \end{tabular}
    \vspace{-1em}
\end{table}

%% file: figures/visual_comparison.tex
\begin{figure*}[t]
    \setlength{\tabcolsep}{1pt}
    \def\imW{0.155\linewidth}
    \def\imhW{0.075\linewidth}
    \centering
    \begin{tabular}{cccccc}

     \mfigure{\imhW}{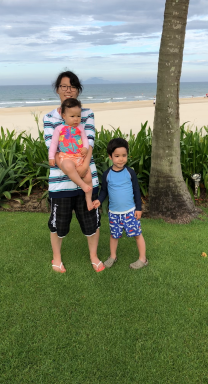}\mfigure{\imhW}{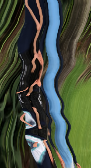} &\mfigure{\imhW}{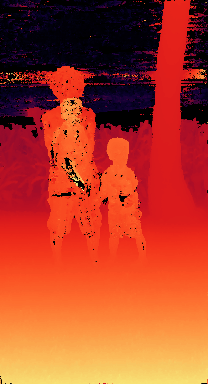}\mfigure{\imhW}{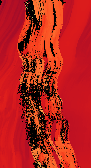} &\mfigure{\imhW}{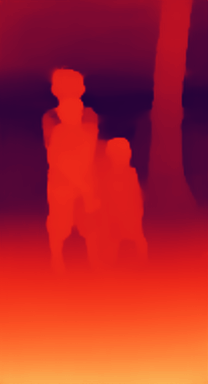}\mfigure{\imhW}{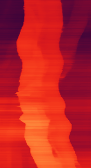} &\mfigure{\imhW}{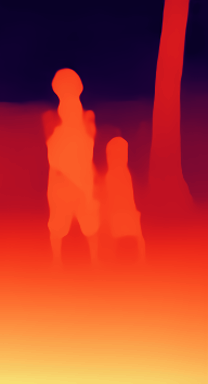}\mfigure{\imhW}{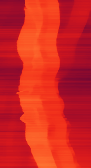} &\mfigure{\imhW}{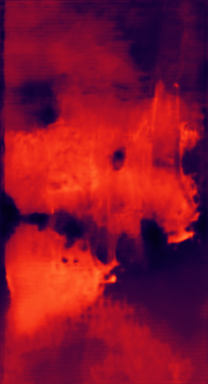}\mfigure{\imhW}{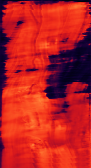} &\mfigure{\imhW}{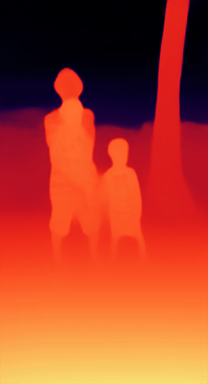}\mfigure{\imhW}{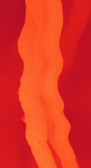}\\
     
    \mfigure{\imhW}{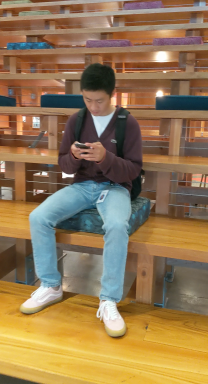}\mfigure{\imhW}{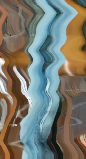} &\mfigure{\imhW}{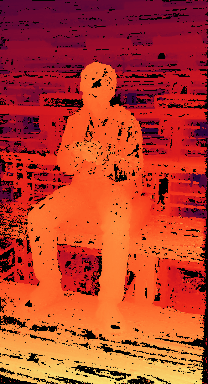}\mfigure{\imhW}{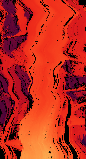} &\mfigure{\imhW}{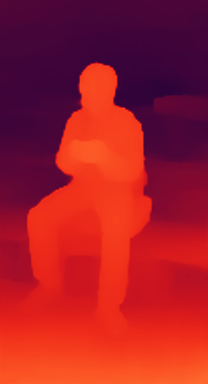}\mfigure{\imhW}{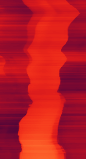} &\mfigure{\imhW}{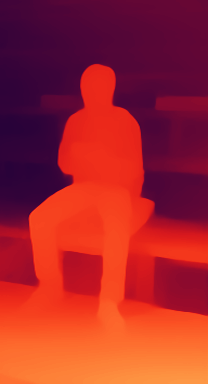}\mfigure{\imhW}{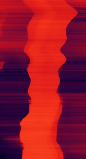} &\mfigure{\imhW}{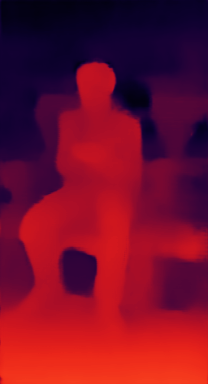}\mfigure{\imhW}{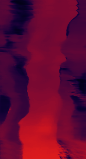} &\mfigure{\imhW}{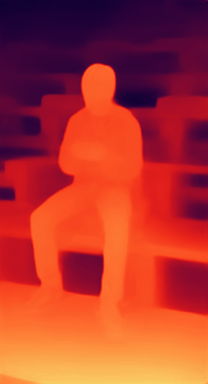}\mfigure{\imhW}{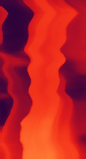}\\
    
    \mfigure{\imhW}{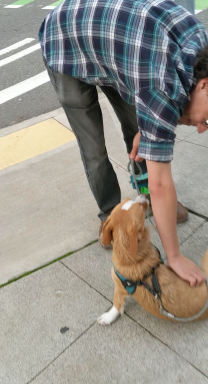}\mfigure{\imhW}{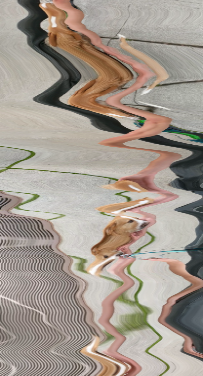} &\mfigure{\imhW}{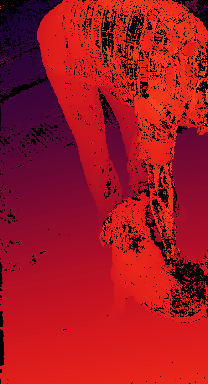}\mfigure{\imhW}{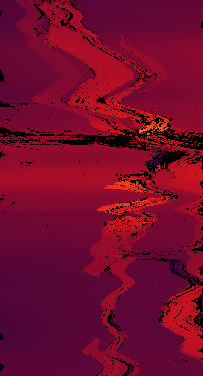} &\mfigure{\imhW}{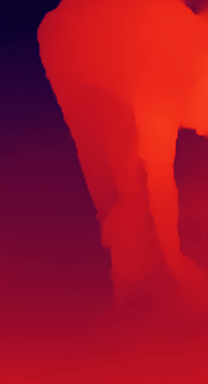}\mfigure{\imhW}{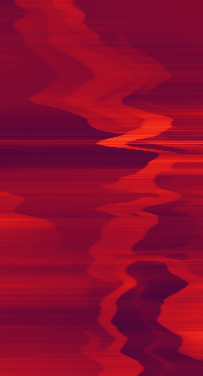} &\mfigure{\imhW}{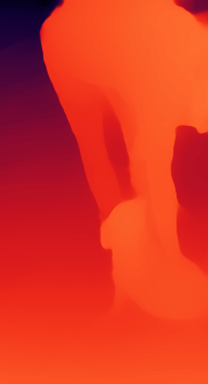}\mfigure{\imhW}{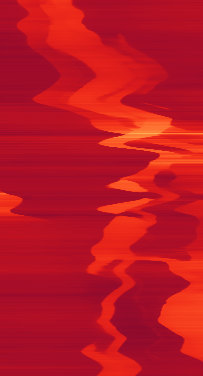} &\mfigure{\imhW}{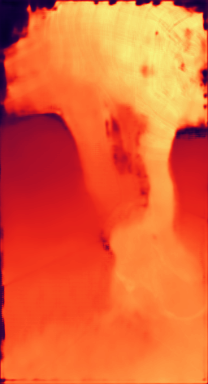}\mfigure{\imhW}{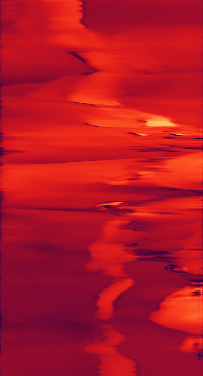} &\mfigure{\imhW}{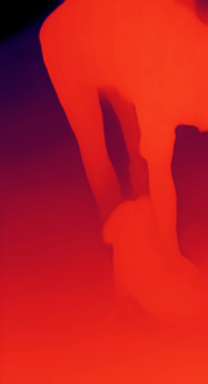}\mfigure{\imhW}{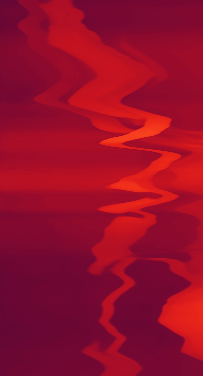}\\

    \mfigure{\imW}{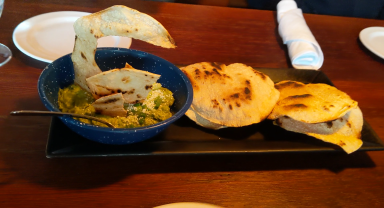} &\mfigure{\imW}{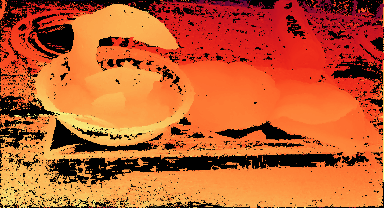} &\mfigure{\imW}{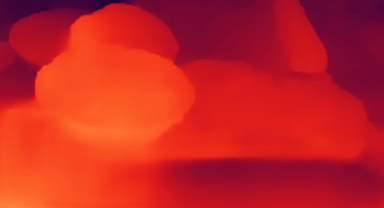} &\mfigure{\imW}{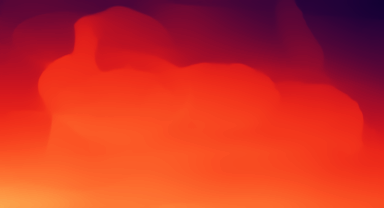} &\mfigure{\imW}{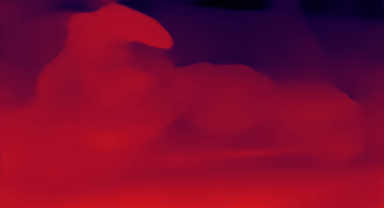} &\mfigure{\imW}{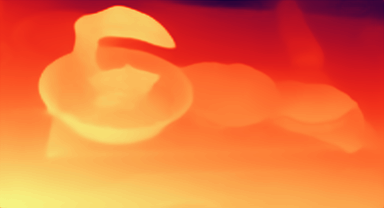}\\
    
    \mfigure{\imW}{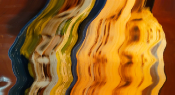} &\mfigure{\imW}{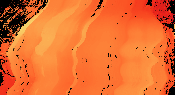} &\mfigure{\imW}{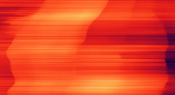} &\mfigure{\imW}{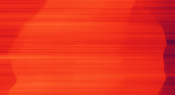} &\mfigure{\imW}{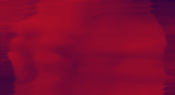} &\mfigure{\imW}{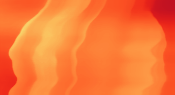}\\
    
    \mfigure{\imW}{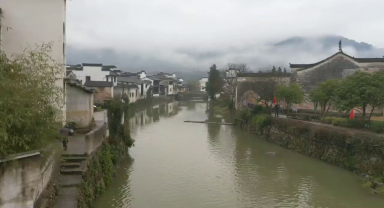} &\mfigure{\imW}{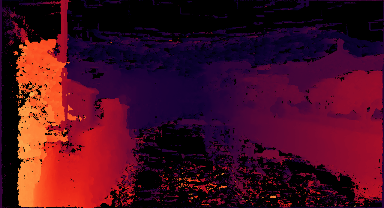} &\mfigure{\imW}{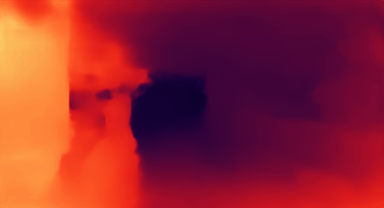} &\mfigure{\imW}{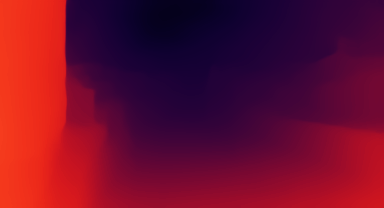} &\mfigure{\imW}{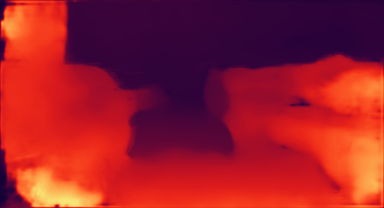} &\mfigure{\imW}{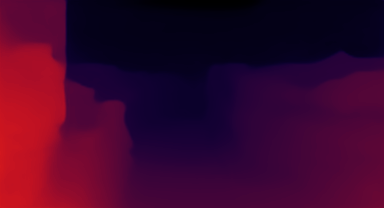}\\
    
    \mfigure{\imW}{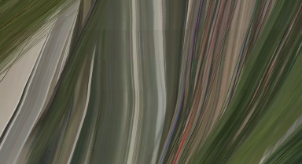} &\mfigure{\imW}{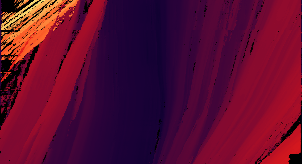} &\mfigure{\imW}{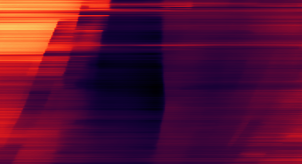} &\mfigure{\imW}{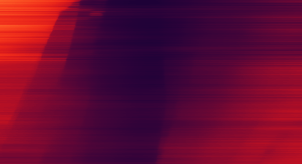} &\mfigure{\imW}{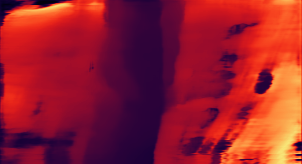} &\mfigure{\imW}{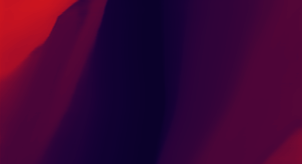} \\
    
    \mpage{\imW}{(a) Input} 
    & \mpage{\imW}{(b) COLMAP} 
     & \mpage{\imW}{(c)Mannequin}  \hfill
    & \mpage{\imW}{(d) MiDaS-v2} 
    & \mpage{\imW}{(e) NeuralRGBD}
    & \mpage{\imW}{(f) Ours}\\
    
    \end{tabular}{}
    \vspace{-3mm}
    \caption{{Visual comparisons with the state-of-the-arts}. Our method produces depth, geometrically consistent, and flicker-free depth estimation from casually captured videos by a hand-held cellphone camera.
    The first image in each pair is a sample frame, while the second is a scanline slice
    through the spatio-temporal volume (either color video or video depth).
    }
    \label{fig:visual-comparison}
    \vspace{-1em}
\end{figure*}

%% file: figures/visual_ablation.tex
\begin{figure*}[t]
    \setlength{\tabcolsep}{1pt}
    \def\imW{0.155\linewidth}
    \begin{tabular}{c|cc|cc|c}
        \mfigure{\imW}{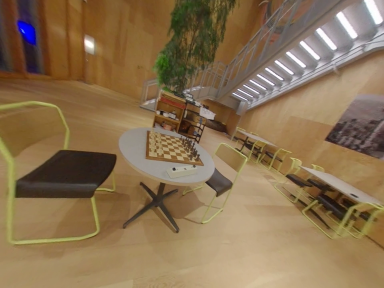}
        & \mfigure{\imW}{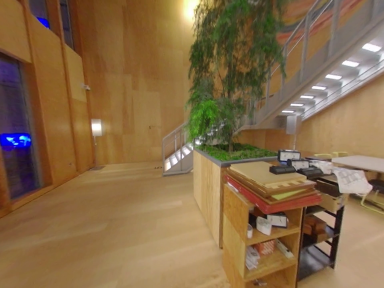} & \mfigure{\imW}{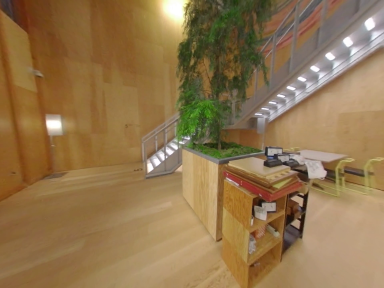} 
        & \mfigure{\imW}{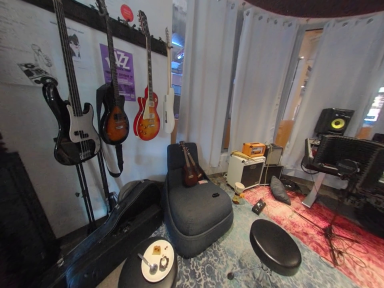} &\mfigure{\imW}{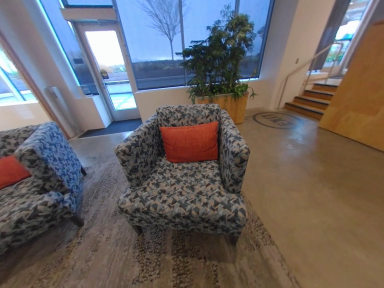} 
        & \mfigure{\imW}{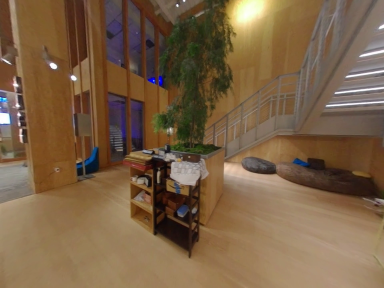}\\
        \mfigure{\imW}{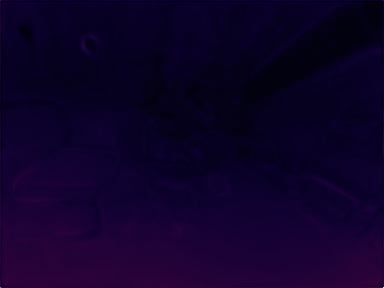}
        &  \mfigure{\imW}{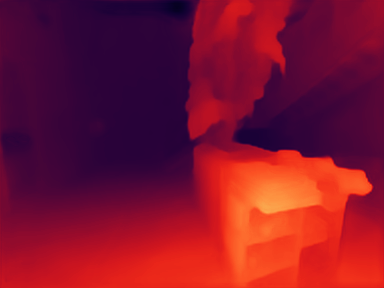}  & \mfigure{\imW}{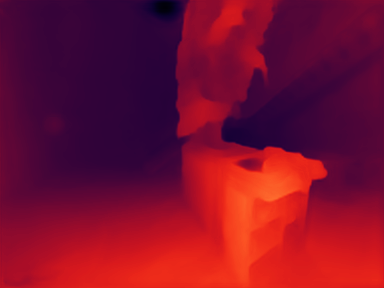}
        &\mfigure{\imW}{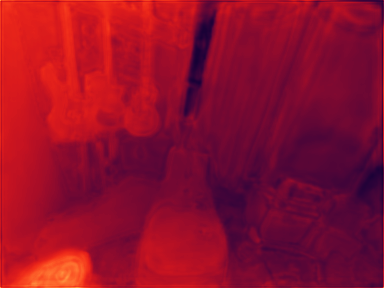} & \mfigure{\imW}{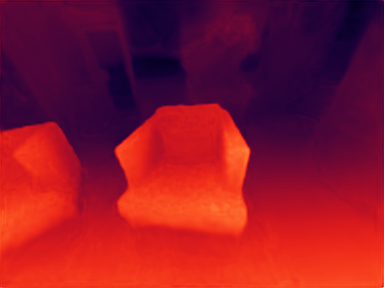} 
        & \mfigure{\imW}{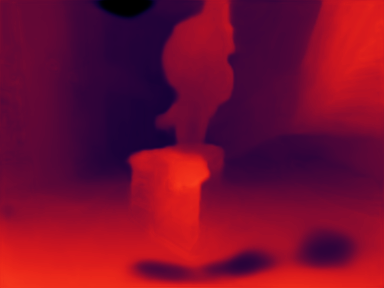} \\
        \mfigure{\imW}{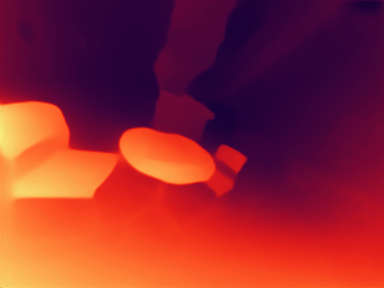}
        &\mfigure{\imW}{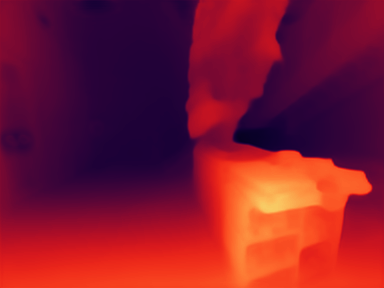} & \mfigure{\imW}{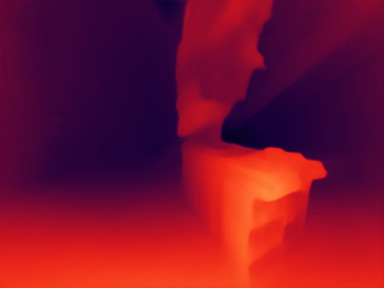}
        &\mfigure{\imW}{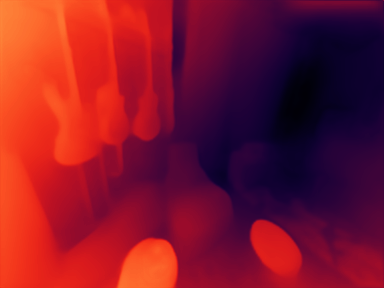} & \mfigure{\imW}{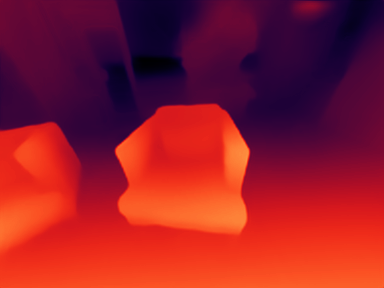}
        & \mfigure{\imW}{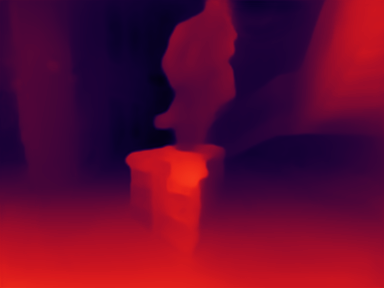}\\
        w/o spatial loss & 
        \multicolumn{2}{c|}{w/o disparity loss} &   \multicolumn{2}{c|}{w/o scale calibration} &
        \rev{w/o overlap test}\\
    \end{tabular}
    \centering
    \vspace{-1em}
    \caption{
    {Contribution of our design choices to the results.} 
    (\emph{Top}): Sample frame from input videos. 
    (\emph{Middle}): corresponding ablation results. Bottom: result with full pipeline. Without spatial loss, there is no constraint for what the depth should be. We end up losing all the structure and it fails. Without disparity loss, depth can get sharper but also more flicker. Without scale calibration, we often observe degraded depth with blurrier depth discontinuities. Without overlap test, erroneous flow causes wrong depth.}
    \label{fig:visual-ablation}
\end{figure*}

%% file: tables/ablation-study.tex
\begin{table}[t]
    \setlength{\tabcolsep}{2pt}
    \centering
    \caption{Ablation study. The quantitative evaluation highlights the importance of our method design choices. 
    }
    \label{tab:ablation}
    \begin{tabular}{lccc}
    \toprule
        & $E_s (\%)$ $\downarrow$ & $E_{d} (\%)$ $\downarrow$ & $E_p$ $\downarrow$ \\
        \midrule
    Ours w/o scale calibration & 0.93 &3.37 & \textbf{9.99} \\
    Ours w/o disparity loss & 0.76 & 3.30 & \textbf{9.99} \\
    Ours w/o overlap test & 0.51 &2.49 & 13.20 \\
    Ours & \textbf{0.44} & \textbf{2.12} & 10.08\\ 
    \bottomrule
    \end{tabular}
\end{table}

%% file: figures/visual_long_term_ablation.tex
\begin{figure}[t]
    \setlength{\tabcolsep}{1pt}
    \def\imW{0.26\linewidth}
    \def\imWW{0.4\linewidth}
    \centering
    \begin{tabular}{cccc}
    \multirow[c]{2}{*}[3.1em]{\mfigure{\imWW}{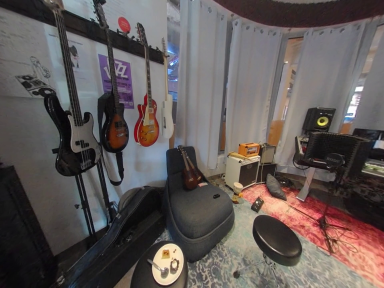}} & \rotatebox{90}{\small{w/o disparity}} & \mfigure{\imW}{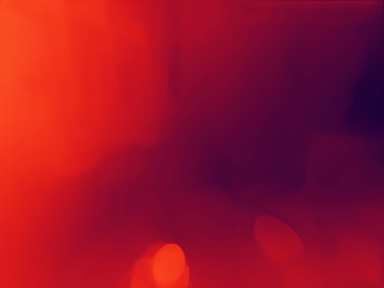} & \mfigure{\imW}{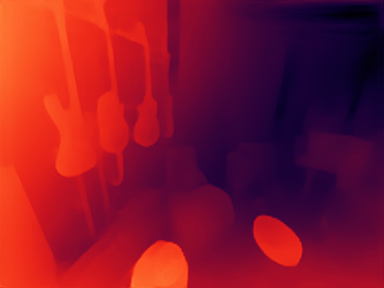} \\
        & \rotatebox{90}{\small{with disparity}} & \mfigure{\imW}{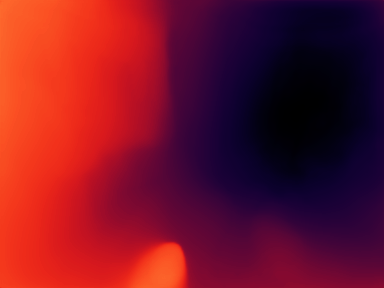}  &\mfigure{\imW}{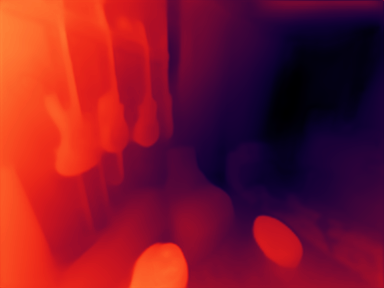}\\
    \midrule
    \multirow[c]{2}{*}[3.1em]{\mfigure{\imWW}{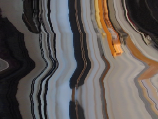}} & \rotatebox{90}{\small{w/o disparity}} & \mfigure{\imW}{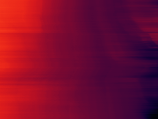} & \mfigure{\imW}{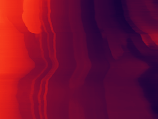}\\
        & \rotatebox{90}{\small{with disparity}} &\mfigure{\imW}{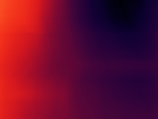} & \mfigure{\imW}{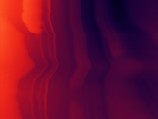}\\
    & & Consecutive & Long-term \\
    \end{tabular}
    \vspace{-3mm}
    \caption{
    Analysis of the effects of using long-term temporal constraints and the disparity loss. Please see the supplementary for video comparisons.
    }
    \label{fig:ablate-long-term}
    \vspace{-1em}
\end{figure}

%% file: figures/quantitative_kitti.tex
\begin{figure}[t]
    \setlength{\tabcolsep}{1pt}
    \def\imW{0.49\linewidth}
    \centering
    \begin{tabular}{cc}
    \mfigure{\imW}{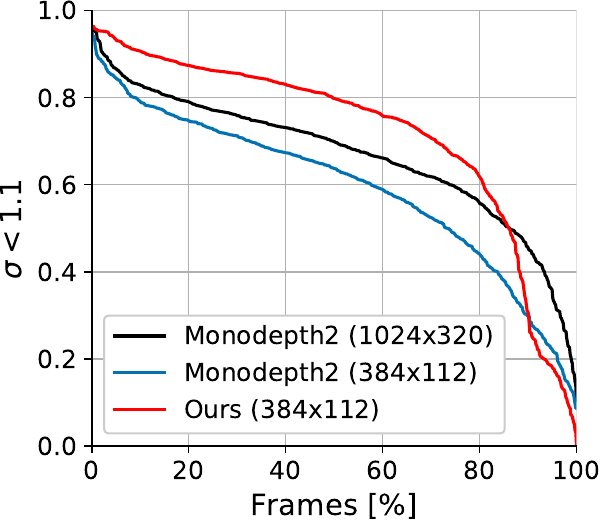} 
    &\mfigure{\imW}{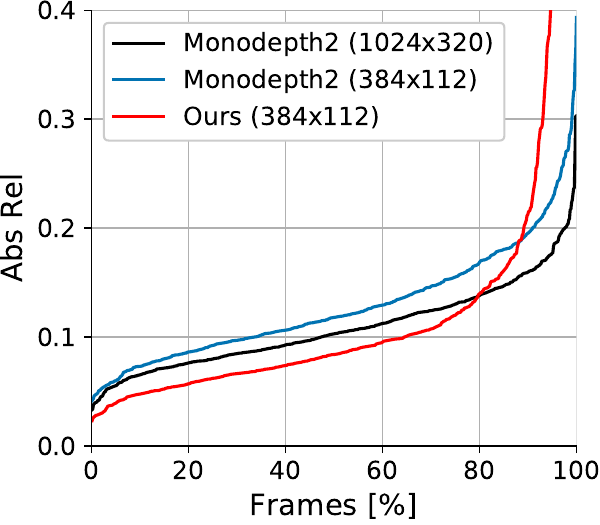}\\
    \end{tabular}
    \vspace{-5mm}
    \caption{\rev{Quantitative comparison before and after fine-tuning Monodepth2 on KITTI. We plot the each metric over all the test frames sorted by their values. After fine-tuning, we have more outliers due to extreme dynamic motion or failure in camera pose estimation, but achieve improved results for more than $80\%$ of the frames.}}
    \label{fig:quantitative-kitti}
    \vspace{-1em}
\end{figure}

%% file: tables/quantitative_scannet.tex
\begin{table}[t!]
    \setlength{\tabcolsep}{2pt}
    \centering
    \caption{
    \rev{
    Quantitative comparison on the ScanNet dataset \cite{dai2017scannet} using the test split provided by Tang and Tan~\shortcite{tang2018ba}.
    }
    }
    \label{tab:quan_ScanNet}
    \vspace{-3mm}
    \resizebox{\linewidth}{!}{
    \begin{tabular}{lccccc}
	\toprule
	    & \multicolumn{4}{c}{Error metric $\downarrow$} \\
	    \cmidrule(lr){2-6} 
	  	& 	Abs Rel	& 	Sq Rel	& 	RMSE	& 	RMSE log	& 	Sc Inv\\
	\midrule 
    DeMoN \shortcite{ummenhofer2017demon}	& 0.231	& 0.520	& 0.761	& 0.289	& 0.284\\
    BA-Net \shortcite{tang2018ba}	& 0.161	& 0.092	& 0.346	& 0.214	& 0.184\\
    DeepV2D (NYU) \shortcite{teed2020deepv2d}	& 0.080	& \ul{0.018}	& 0.223	& 0.109	& 0.105\\
    DeepV2D (ScanNet) \shortcite{teed2020deepv2d} & \textbf{0.057}	& \textbf{0.010}	& \textbf{0.168}	& \textbf{0.080}	& \textbf{0.077}\\
	MiDaS-v2 \shortcite{li2019learning}	&0.208	& 0.318	& 0.742	& 0.246	& 0.239\\
	Ours	&	\ul{0.073}	& 	0.037	& 	\ul{0.217}	& 	\ul{0.105}	& 	\ul{0.103}\\
	\bottomrule 
    \end{tabular}
    }
    \vspace{-1em}
\end{table}

%% file: tables/quantitative_tum.tex
\begin{table*}[t]
    \setlength{\tabcolsep}{2pt}
    \centering
    \caption{
    \rev{
    Quantitative comparison on the TUM-RGBD dataset (3D Object Reconstruction category) \cite{sturm12iros} in the disparity space. We report the averaged results over 11 video sequences.
    }
    }
    \label{tab:quan_TUM}
    \begin{tabular}{llcccccccc}
    \toprule
        & 
        & \multicolumn{4}{c}{Error metric $\downarrow$} & \multicolumn{3}{c}{Accuracy metric $\uparrow$} \\
        \cmidrule(lr){3-6}  \cmidrule(lr){7-9}
		& & 	Abs Rel	& 	Sq Rel	& 	RMSE	& 	RMSE log	& 	$\sigma < 1.25$	& 	$\sigma < 1.25^2$	& 	$\sigma < 1.25^3$\\
	\midrule 
	\multirow{2}{*}{Single-frame} & 
	Mannequin \shortcite{li2019learning}	&0.306	& 0.101	& 0.244	& 0.385	& 0.569	& 0.772	& 0.885\\
	    & MiDaS-v2 \shortcite{li2019learning}	& \ul{0.220}	& \ul{0.061}	& \ul{0.187}	& \ul{0.292}	& \ul{0.665}	& \ul{0.861}	& \ul{0.945} \\
	\midrule
	\multirow{3}{*}{Multi-frame} & 
	WSVD \shortcite{wang2019web}	& 0.281	& 0.083	& 0.228	& 0.365	& 0.551	& 0.794	& 0.905\\
	    & NeuralRGBD \shortcite{liu2019neural}	&0.615	& 0.365	& 0.392	& 0.661	& 0.361	& 0.571	& 0.710\\
	    & Ours	&	\textbf{0.144}	& 	\textbf{0.036}	& 	\textbf{0.144}	& 	\textbf{0.211}	& 	\textbf{0.785}	& 	\textbf{0.934}	& 	\textbf{0.979}\\
	\bottomrule 
    \end{tabular}
\end{table*}


%% file: tables/quantitative_kitti.tex
\begin{table*}[t]
    \setlength{\tabcolsep}{2pt}
    \centering
    \caption{
    \rev{
    Quantitative comparisons with existing methods on the KITTI benchmark dataset using the Eigne split. (\emph{Top}): methods that produce full resolution ($1024 \times 320$) depth maps. (\emph{Bottom}): methods that produce low-resolution ($384 \times 112$) depth maps. 
    Note that for fair comparison, we align the depth results from all the compared methods with per-image median ground truth scaling. 
    Therefore, our reported numbers for Monodepth2 ($1024\times 320$)~\shortcite{godard2019digging} differ slightly from those in their paper where they use a constant scale for alignment.
    } 
    }
    \label{tab:quan_KITTI}
    \begin{tabular}{lccccccc}
	\toprule
	    & \multicolumn{4}{c}{Error metric $\downarrow$} & \multicolumn{3}{c}{Accuracy metric $\uparrow$} \\
        \cmidrule(lr){2-5}  \cmidrule(lr){6-8}
		& 	Abs Rel	& 	Sq Rel	& 	RMSE	& 	RMSE log	& 	$\sigma < 1.25$	& 	$\sigma < 1.25^2$	& 	$\sigma < 1.25^3$\\
	\midrule 
    Zhou \shortcite{zhou2017unsupervised} & 0.183	& 1.595	& 6.709	& 0.270	& 0.734	& 0.902	& 0.959\\
    GeoNet \shortcite{yin2018geonet}	& 0.149	& 1.060	& 5.567	& 0.226	& 0.796	& 0.935	& 0.975\\
    DF-Net \shortcite{zou2018df}	& 0.150	& 1.124	& 5.507	& 0.223	& 0.806	& 0.933	& 0.973\\
    Struct2depth \shortcite{casser2019depth} & 0.109 & 0.825 & 4.750 & \ul{0.187} & 0.874 & \ul{0.958} & \textbf{0.983} \\
    GLNet \shortcite{chen2019self} & \textbf{0.099} & \textbf{0.796} & \ul{4.743} & \textbf{0.186} & \ul{0.884} & 0.955 & 0.979 \\
	Monodepth2 ($1024\times 320$) \shortcite{godard2019digging}	&	\ul{0.108}	& 	\ul{0.806}	& 	\textbf{4.606}	& 	\ul{0.187}	& 	\textbf{0.887}	& \textbf{0.962}  & \ul{0.981}\\
	\midrule
	Monodepth2 ($384\times 112$) \shortcite{godard2019digging}	&0.128	& 1.040	& 5.216	& 0.207	& 0.849	& 0.951	& 0.978\\
	Ours ($384\times 112$)	&0.130	& 2.086	& 4.876	& 0.205	& 0.878	& 0.946	& 0.970\\
	\bottomrule 
    \end{tabular}
\end{table*}

%% file: figures/effects.tex
\setlength\height{5.4cm}%
\newlength\neggap%
\setlength\neggap{-1mm}%
\begin{figure*}%
\centering%
\jsubfig{\includegraphics[height=\height, trim=0 100 0 200, clip]{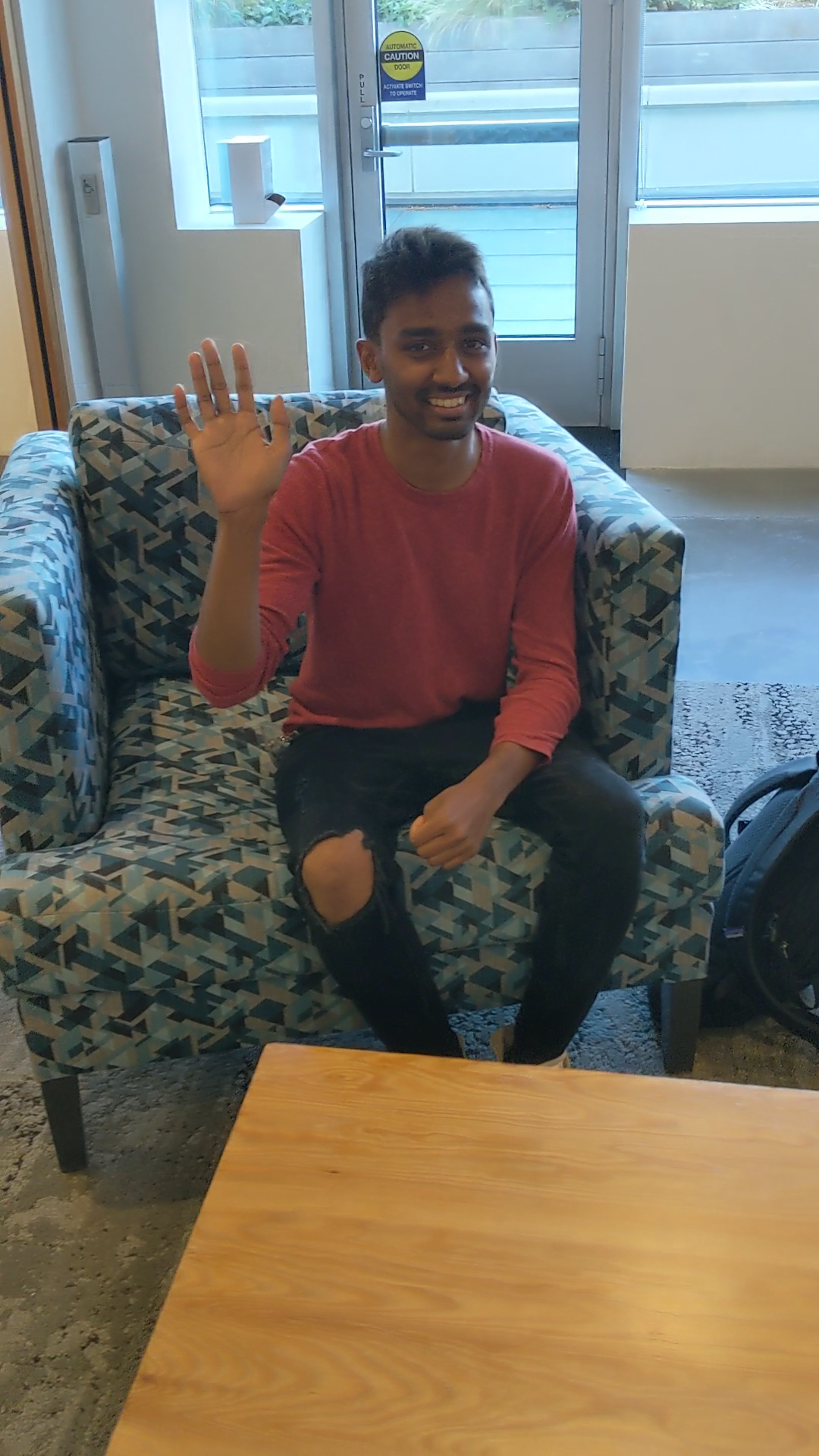}}{\vspace{\neggap}%
  Input video}%
\hfill%
\jsubfig{\includegraphics[height=\height, trim=0 50 0 100, clip]{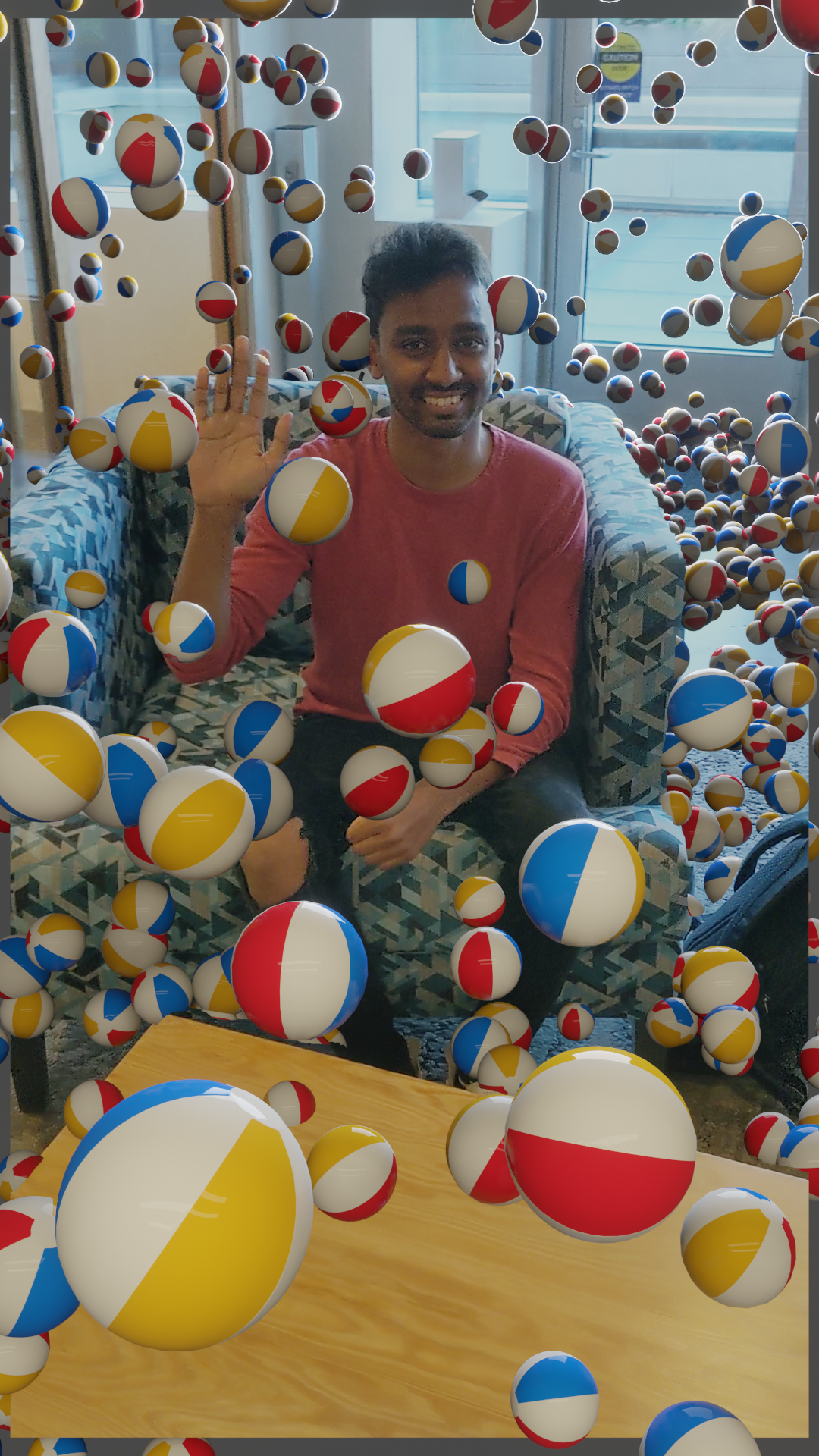}}{\vspace{\neggap}%
  Bouncing balls}%
\hfill%
\jsubfig{\includegraphics[height=\height, trim=0 50 0 100, clip]{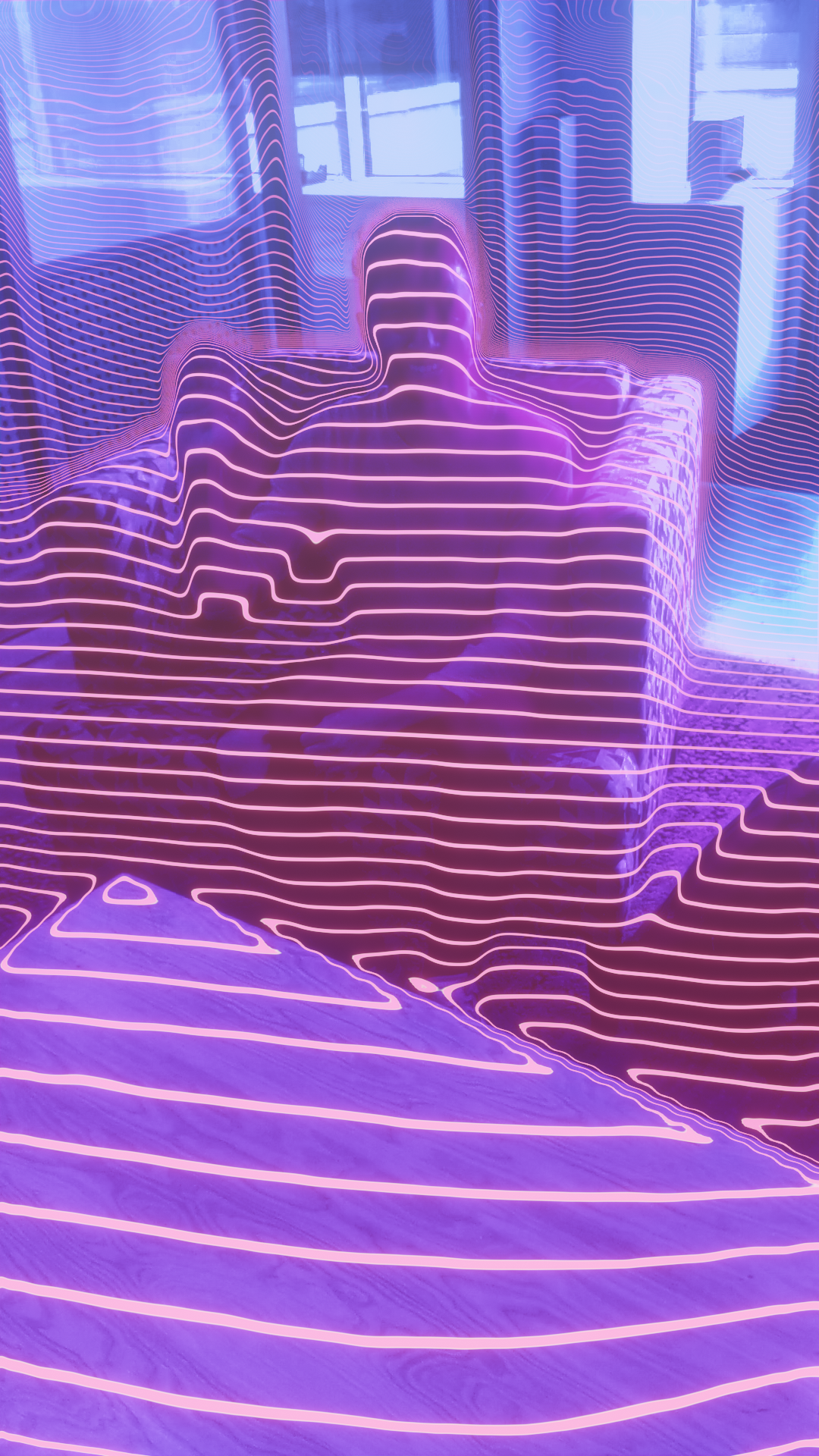}}{\vspace{\neggap}%
  Disco}%
\hfill%
\jsubfig{\includegraphics[height=\height, trim=0 50 0 100, clip]{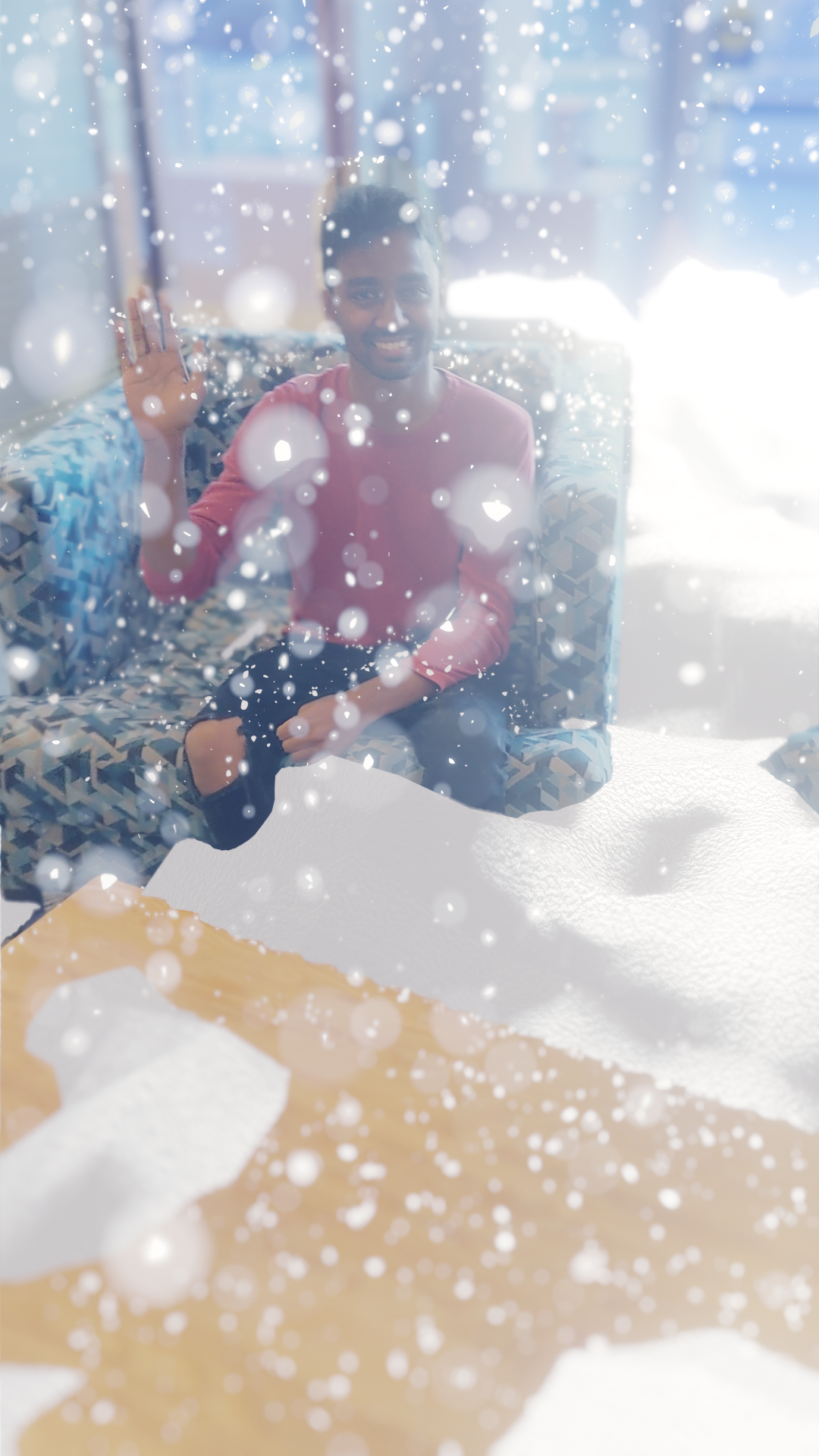}}{\vspace{\neggap}%
  Snow}%
\hfill%
\jsubfig{\includegraphics[height=\height, trim=0 50 0 100, clip]{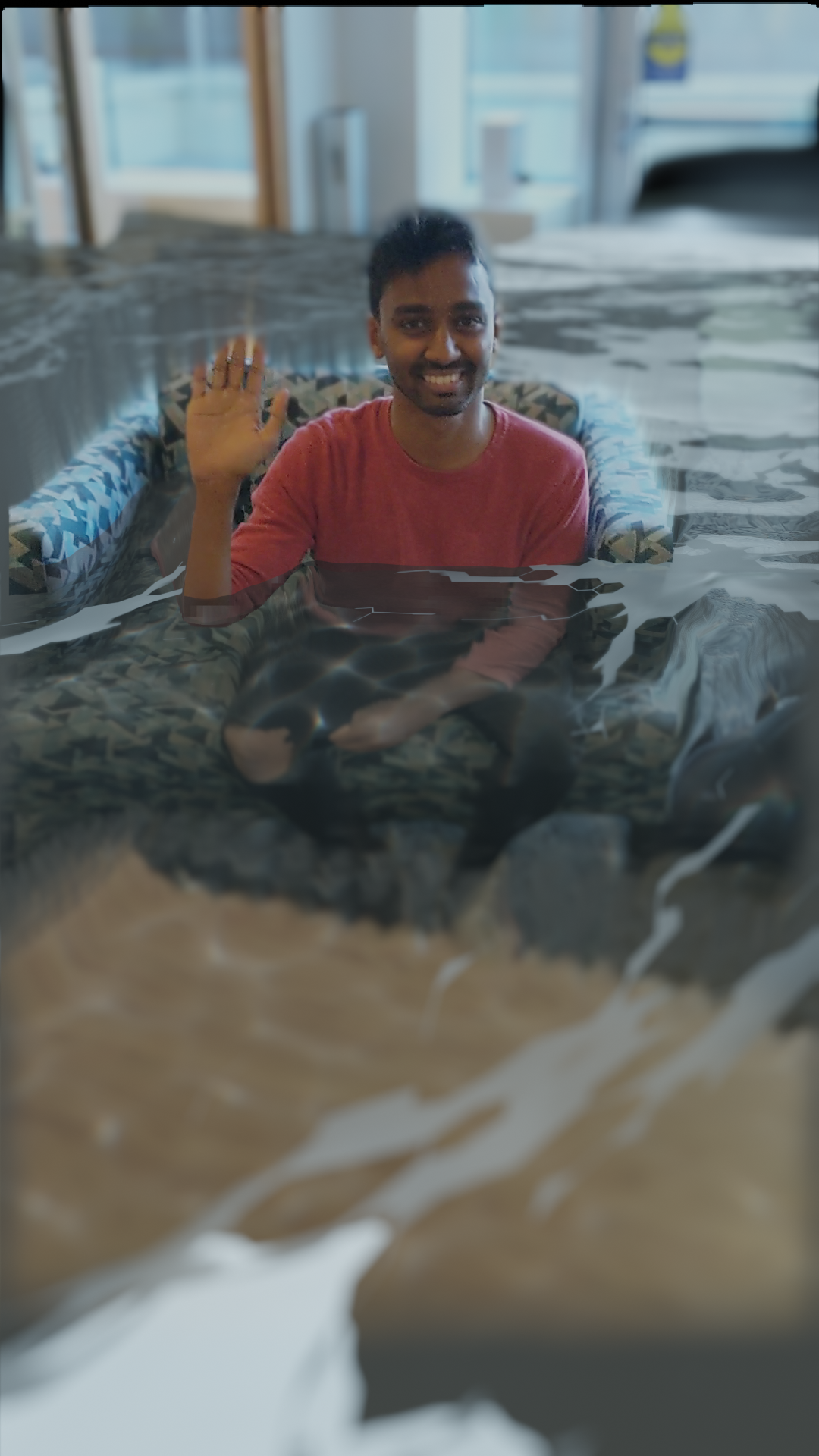}}{\vspace{\neggap}%
  Water}%
\vspace{-3mm}
\def\captiontext{Our consistent depth estimation enables a wide range of fully-automated video-based visual effects. We refer the readers to the supplementary video.}
\caption{\captiontext}
\Description[Short description]{\captiontext}
\label{fig:effects}
\end{figure*}
\undef\height
\undef\neggap

%% file: tex/conclusions.tex
\section{Conclusions}

We have presented a simple yet effective method for estimating \emph{consistent} depth from a monocular video. 
Our idea is to leverage geometric constraints extracted using conventional multi-view reconstruction methods and use them to fine-tune a pre-trained single-image depth estimation network. 
Using our test-time fine-tuning strategy, our network learns to produce geometrically consistent depth estimates across entire video. 
We conduct extensive quantitative and qualitative evaluation.
Our results show that our method compares favorably against several state-of-the-art depth estimation algorithms. 
Our consistent video depth estimation enables compelling video-based visual effects. 




%% file: tex/acknowledge.tex
\begin{acks}
We would like to thank Patricio Gonzales Vivo, Dionisio Blanco, and Ocean Quigley for creating the artistic effects in the accompanying video. We also thank True Price for his practical and insightful advice on reconstruction and Ayush Saraf for his suggestions in engineering.
\end{acks}